\documentclass[letterpaper]{article} 
\usepackage{aaai24}  
\usepackage{times}  
\usepackage{helvet}  
\usepackage{courier}  
\usepackage[hyphens]{url}  
\usepackage{graphicx} 
\usepackage{natbib}  
\usepackage{caption} 
\usepackage{algorithm}
\usepackage{algorithmic}

\usepackage{newfloat}
\usepackage{listings}
\usepackage{xcolor}
\newcommand{\mj}[1]{\textcolor{black}{#1}}

\newcommand{\kt}[1]{\textcolor{black}{#1}}

\usepackage{latexsym}
\usepackage[T1]{fontenc}
\usepackage[utf8]{inputenc}
\usepackage{microtype}
\usepackage{inconsolata}
\usepackage{booktabs}
\usepackage{kotex}
\usepackage{multirow}
\usepackage{amsmath, amssymb}
\usepackage{tabularx}
\usepackage{caption}
\usepackage{subcaption}
\usepackage{adjustbox}
\DeclareMathOperator*{\argmax}{arg\,max} 
\usepackage{pifont}
\newcommand{\cmark}{\ding{51}}%
\newcommand{\xmark}{\ding{55}}%

\DeclareCaptionStyle{ruled}{labelfont=normalfont,labelsep=colon,strut=off} 
\floatstyle{ruled}
\newfloat{listing}{tb}{lst}{}
\floatname{listing}{Listing}
\title{BOK-VQA: Bilingual outside Knowledge-Based Visual Question Answering via Graph Representation Pretraining}
\author{
    MinJun Kim\textsuperscript{\rm 1}\equalcontrib,
    SeungWoo Song\textsuperscript{\rm 1}\equalcontrib,
    YouHan Lee\textsuperscript{\rm 2},
    Haneol Jang\textsuperscript{\rm 1},
    KyungTae Lim\textsuperscript{\rm 3}\thanks{Corresponding author.}
}
\affiliations{
    \textsuperscript{\rm 1} Hanbat National University\\ 
    \textsuperscript{\rm 2} Kakao Brain\\ 
    \textsuperscript{\rm3} Seoul National University of Science and Technology\\

    mjkmain@seoultech.ac.kr, sswoo@seoultech.ac.kr, \\youhan.lee@kakaobrain.com,
    hejang@hanbat.ac.kr, ktlim@seoultech.ac.kr
}

\usepackage{bibentry}

\begin{document}

\maketitle

\begin{abstract}
The current research direction in generative models, such as the recently developed GPT4, aims to find relevant knowledge information for multimodal and multilingual inputs to provide answers. Under these research circumstances, the demand for multilingual evaluation of visual question answering (VQA) tasks, a representative task of multimodal systems, has increased. 
Accordingly, we propose a bilingual outside-knowledge VQA (BOK-VQA) dataset in this study that can be extended to multilingualism. The proposed data include 17K images, 17K question-answer pairs for both Korean and English and 280K instances of knowledge information related to question-answer content. We also present a framework that can effectively inject knowledge information into a VQA system by pretraining the knowledge information of BOK-VQA data in the form of graph embeddings. Finally, through in-depth analysis, we demonstrated the actual effect of the knowledge information contained in the constructed training data on VQA.
\end{abstract}

\section{Introduction}\label{sec:introduction}
Visual question answering (VQA) \cite{antol2015vqa} refers to a system that provides answers to natural language questions related to input images. In recent years, an extended form of VQA, called knowledge based VQA (KB-VQA), has developed that utilizes external knowledge sources, such as Wikipedia or knowledge graphs (KGs), to enable higher-level reasoning to answer complex questions. This knowledge-based approach in VQA leverages various types of information, including biographical (K-VQA) \cite{K-VQA}, common sense (FVQA) \cite{wang2017fvqa}, (KB-VQA) \cite{KB-VQA}, Wikipedia abstracts (OK-VQA) \cite{schwenk2022okvqa}, and scene graphs (GQA) \cite{hudson2019gqa}.

However, despite these advancements, the existing versions of KB-VQA encounter various challenges. First, owing to the cost of construction, the size of training data is either relatively small (FVQA and KB-VQA) or generated automatically based on metadata (GQA). 
Second, most datasets are constructed predominantly in English. Given these language biases, a recent approach for addressing this issue introduced xGQA \cite{pfeiffer2022xgqa}, a multilingual VQA evaluation dataset for a few-shot evaluation, however, xGQA is limited by the lack of sufficient training data.

Further, KB-VQA encounters challenges in the construction of training data for less-resourced languages owing to the need for accompanying information (e.g., Wikipedia). The consideration of this issue gives rise to the major research question of this study, i.e., can we leverage the rich external knowledge of highly-resourced languages to construct KB-VQA training data for less-resourced languages? We attempted to address this question in this study. Considering that it is not necessary to rely on a specific language if utilizing the knowledge representation of highly-resourced languages in the form of entity embeddings, we proposed a bilingual dataset and method for representing knowledge through KG Embeddings (KGEs). 

The proposed dataset possessed the following characteristics: (1) We constructed a relatively large-scale training dataset using 17K Korean-English question-answer pairs and 280K instances of information. (2) It is compiled with the structure of bilingual outside-knowledge VQA (BOK-VQA), which utilizes knowledge base (KB) information from English, a representative high-resource language, in conjunction with other languages.

KGs, which typically consist of object and relation information, can be used to represent complex relationships. Thus, KGE methods \cite{rescal,transE,kge, complEX, ConvE, rotatE} have been proposed as techniques for compressing and representing data while preserving the structure and information of KGs. 
This leads us to the final question of this study, i.e., how can knowledge embeddings be efficiently fused into VQA? \kt{For that, we propose a graph-embedded learning-based VQA (\textsc{gel-vqa}) model. This model acquires knowledge information from KGs through triple prediction and subsequently leverages this information to address VQA problems.}
The proposed multitasking model is designed to minimize the loss of information across different tasks by predicting the information required to solve VQA questions through triple prediction and concurrently conducting VQA training.

\kt{Through in-depth experiments, we} revealed that the performance of VQA varied widely from 21.16 to 55.48 depending on the application of KGE and triple prediction. This indicates that the proposed dataset represents a problem that can be solved by utilizing KB information, and KG-based methods have a substantial impact. In addition, to analyze the impact of KGE methods on the \textsc{gel-vqa} model, we evaluate the performance of the proposed model based on different KGE methods. The contributions of this study are given below:
\begin{itemize}
\item We propose a multilingual extensible KB-VQA training dataset named BOK-VQA.
\item We propose a language-independent \textsc{gel-vqa} training framework for KG information.
\item Through in-depth analysis, we prove the actual effect of language-independent KGs on BOK-VQA.
\end{itemize}
\section{Related Works}\label{sec:related-work}
VQA was first introduced in 2015 by \citeauthor{antol2015vqa}, and the dataset comprised questions that could be answered with only a shallow understanding of images.

Moving beyond simple questions, VQA that used more complex external knowledge was scarcely explored until \citeauthor{wang2017fvqa} proposed the FVQA dataset. FVQA employed DBpedia, ConceptNet, and Webchild to incorporate common sense and object-related knowledge into question-answering. However, despite the use of KGs, the questions mainly focused on object recognition in images, and object-related question-answering was similar to traditional knowledge-based question-answering. To address these issues, outside knowledge base VQA (OK-VQA) \cite{schwenk2022okvqa}, which is less reliant on KBs, was proposed. OK-VQA includes information based on web texts and other sources, offering greater flexibility, but at the cost of lower knowledge utilization. Subsequently, scene graph-based VQA (GQA) \kt{\cite{hudson2019gqa} was introduced in 2019, which utilized scene graph information from the Visual Genome dataset \cite{visualgenome} to automatically generate a large volume of VQA datasets.} 
Efforts to expand VQA into multilingual settings began in 2021. Because most VQA-related datasets are based on monolingual (English) sources, there is a lack of research on VQA for low-resource languages. Therefore, xGQA was proposed to evaluate multilingual VQA. However, since it was mainly constructed for evaluation in various languages through few-shot learning rather than as training data, using it for training purposes is challenging.

Additionally, VQA datasets have been proposed for domain-specific purposes, such as MovieQA \cite{tapaswi2016movieqa}, TVQA \cite{lei2018tvqa}, DramaQA \cite{choi2021dramaqa}, and ActivityNet-QA \cite{yu2019activitynet}. MovieQA is a video-based VQA dataset comprising 15,000 question-answer pairs obtained from captures, plots, lyrics, and scripts from 400 movies. TVQA specializes in six TV show domains and contains 152,000 question-answer pairs from 925 episodes. DramaQA is a question-answering dataset for the Korean drama ``Another Miss Oh'' characterized by its hierarchical structure based on question difficulty. Finally, ActivityNet-QA is an action-oriented VQA dataset with 58,000 question-answer pairs extracted from 5,800 videos. 
\section{Proposed BOK-VQA Dataset}
\label{sec:dataset}
\begin{table*}[!t]
\centering
\small
\begin{tabular}{l>{\centering\arraybackslash}p{3cm}ccc>{\centering\arraybackslash}p{3cm}}
\toprule
    Criterion                      & FVQA             & OK-VQA            & KB-VQA        &KVQA           &  BOK-VQA(Ours) \\
\midrule 
    Number of images               & 2,190            & 14,031            & 700           & 24,602        & 17,836\\
    Number of question-answer pairs & 5,826            & 14,055            & 2,402         & 183,007       & 17,836\\
    Average question length(words) & 9.5              & 8.1               & 6.8           & 10.1          & 10.8 \\
    Number of questions per image  & 2.66             & 1.00              & 3.43          & 7.44          & 1.00  \\
    Number of knowledge sources     & 3                & 1                 & 1             & 1             & 2  \\
    Knowledge source               & DBpedia +ConceptNet +Webchild & Wikipedia & DBpedia & Wikidata       & DBpedia      +ConceptNet\\
    Multilinguality                & \xmark           & \xmark            & \xmark        & \xmark        & \cmark  \\
    
\bottomrule 
\end{tabular}
\caption{Key statistics for KB-VQA dataset.}
\label{tab:kbvqa-dataset-statistics}
\end{table*}

\begin{figure}[t]
    \includegraphics[width=\linewidth]{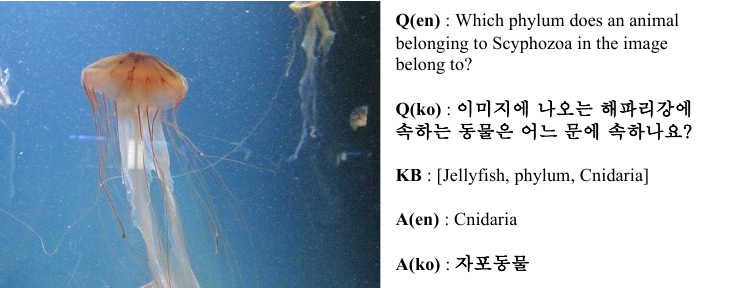}
    \caption{Example of BOK-VQA sample}
    \label{fig:fig1}
\end{figure}

In this study, we propose a BOK-VQA dataset comprising 17,836 samples and 282,533 knowledge triples. Each sample contained of an image, question, answer, and $k$ external knowledge IDs that are necessary to solve a question. Figure~\ref{fig:fig1} presents an example of the BOK-VQA dataset. It is challenging to answer the question, ``Which phylum does an animal belonging to Scyphozoa in the image belong to?'' without using external knowledge. The answer accuracy can be enhanced using external knowledge given in the form of a triple ``[`Jellyfish', `phylum', `Cnidaria']'' with the image-question.

Considering multilingual extensibility, how should knowledge information in KB-VQA be efficiently structured? Data with a unified ontological system for objects expressed in different languages can be advantageous in this case. External knowledge that meets these criteria can be linked to data wherein concepts in different languages are connected through URIs. Accordingly, repositories such as ConceptNet \cite{speer2017conceptnet} and DBpedia \cite{auer2007dbpedia, lehmann2015dbpedia}, which were proposed as common sense KBs in FVQA, can be potentially used.

However, because gathering sufficient KG data is not feasible for low-resource languages, we propose a novel method to use only English KGs. It is noteworthy that because entity and relation information are independent of language, well-trained KGEs, which contain knowledge, are language-agnostic. For example, the surface form of English triple $<$Barack\_Obama, Spouse, Michelle\_Obama$>$ and Korean triple $<$버락\_오바마, Spouse, 미셸\_오바마$>$ is presented differently in two distinct languages, but the meaning of entity and relation information $<$head, relation, tail$>$ is entirely identical. Therefore, by using English KGs as embeddings for Korean KB-VQA, the need for Korean KGs can be eliminated. Specifically, in this study, we pretrained the English KGs as embeddings using ConvKB \cite{nguyen2018novel} and used it for multiple languages. Then the only remaining question is how can a Korean VQA question be linked to a related English triple? We address this problem by utilizing a module that predicts related English triple based on an image and a Korean question. For instance, to answer the Korean question ``버락 오바마의 아내는? (Who is Barack Obama's wife?)'', we predicted the English triple $<$Barack\_Obama, Spouse, Michelle\_Obama$>$ using the given image and Korean question. We then extracted its embedding, which was pretrained using English KB data from the predicted triple.

Considering the aforementioned aspects, we assembled 282,533 triple knowledge entries comprising 1,579 objects and 42 relations from English ConceptNet and DBpedia. The selection criteria for the objects and relations were principally based on the 500 objects and 10 relations used in the FVQA dataset. In addition, considering the usage frequency, we incorporated 1,079 objects derived from ImageNet \cite{imagenet} and supplemented 32 additional relations. Table~\ref{tab:kbvqa-dataset-statistics} presents a basic statistical comparison between the proposed BOK-VQA and other KB-VQA datasets. BOK-VQA utilizes a relatively large number of images and composes longer questions \kt{with} a bilingual context. The object list and relation statistical information utilized in BOK-VQA, along with detailed data samples, can be found in Appendix A. 

\kt{A total of 87 skilled annotators created the question-answer data for BOK-VQA. After the creation of the Korean question data, English questions and answers were generated through neural translation. To ensure the quality of the bilingual QA, two professional translators deeply reviewed the translated English data. Based on this review process, high-quality bilingual data were created. However, there may still be a limitation in the standardization of sentence structures because we revised the translated sentences despite the review process. }

In summary, BOK-VQA inherited the data schema of FVQA, with significantly enhanced reliability and scalability. First, in terms of data reliability, the BOK-VQA dataset was constructed by imposing several constraints during the question generation process, ensuring that questions required knowledge information for solving. \kt{For instance, questions like 'Who invented the instrument in the image?' were selected only if three different reviewers deemed appropriate knowledge information necessary to address the question.} Second, in terms of scalability, the question-answer data were constructed as bilingual training data; moreover, because only English is required for KGs, they can be extended to multiple languages. Further, with the availability of object lists, knowledge information related to objects can be easily accessed in every language.
\section{Model Description}\label{sec:model-description}
In this section, we introduce the proposed \textsc{GEL-VQA} model that combines VQA and KGs by converting them into a representational layer. This model focuses on answering visual questions and acquires external information by predicting KG triples. The proposed model broadly contains three parts. 
\kt{First, the \textbf{VQA module} is designed to train on VQA tasks using image and question features. This module serves as the \textsc{baseline} model.} Second, the \textbf{triple prediction module}, a component intended for knowledge acquisition from the KB, is given a pair of images and questions as inputs. This module acquires the information associated with an image by predicting external knowledge contained in triples. Third, the \textbf{KGE module} that offers a method for incorporating KGs into the VQA module; it uses KGE techniques to generate embedding containing relation information of KGs. The embedding values obtained through this module were fed into the proposed model, which enhanced the model's performance.

\subsection{VQA Module}\label{subsec:VQAModule}
\begin{figure}[H]
  \includegraphics[width=\linewidth]{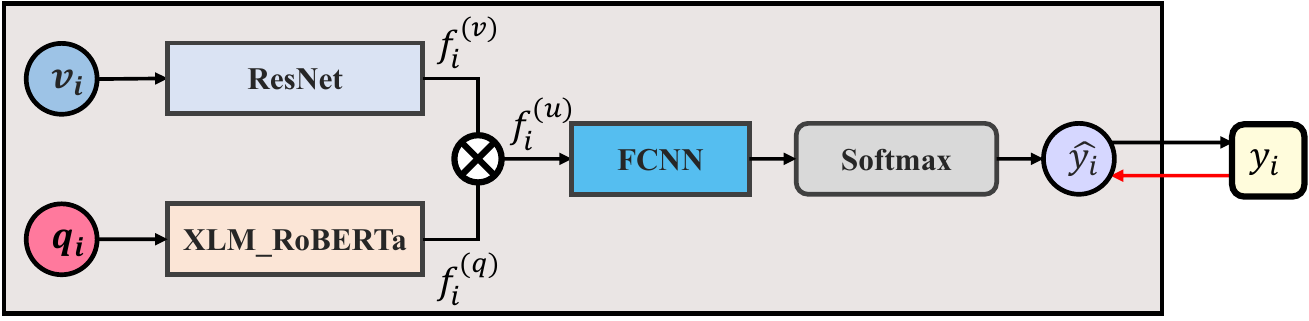}
  \caption{Architecture of VQA module.}
  \label{fig:VQA}
\end{figure}

As presented in Figure ~\ref{fig:VQA}, the VQA module serves as the foundational model and uses images and questions as input data. Given a dataset $D = \{v_i, q_i, y_i\}_{i=1}^N$, where $v_i$ represents the $i$-th image, $q_i$ is the $i$-th question, and $y_i$ is the corresponding answer to $v_i$ and $q_i$, if we denote the VQA module as $P_{vqa}$, our problem can be defined as the following probabilistic model utilizing the method of maximum likelihood:
\begin{equation}
    \theta^{*} = \max_{\theta}\prod_{i=1}^N P_{vqa}(y_i | v_i, q_i; \theta)
\end{equation}
Here, $\theta$ represents the set of model parameters, and $\theta^*$ denotes the optimal parameters of the probabilistic model. $P_{vqa}(y_i | v_i, q_i; \theta)$ represents the probability of accurately predicting the target value $(y_i)$, given the input data $(v_i$, $q_i)$ and the model parameters $(\theta)$. Accordingly, the cross-entropy loss function can be defined as:
\begin{equation}
    \mathcal{L}(\theta) = - \frac{1}{N} \sum_{i=1}^{N} \sum_{j=1}^{C} y_{i, j} \log(P_{vqa}(j | v_i, q_i; \theta))
\end{equation}
Where $y_{i,j}$ is the one-hot encoded representation of the actual class of the $i$-th data sample and $C$ is the total number of classes. In this study, to implement the probability model $(P_{vqa})$ as a multilingual model, we generate a d-dimensional feature vector $f_i^{(q)} \in \mathbb{R}^d$, for the input question $q_i$ through the multilingual-based XLM-RoBERTa \cite{xlmroberta}. For input image $v_i$, we generate a d-dimensional feature vector, $f_i^{(v)} \in \mathbb{R}^d$, via ResNet50 \cite{resnet}. Finally, to combine these two modality features, we create an unified d-dimensional feature vector $f_i^{(u)} \in \mathbb{R}^d$ through element-wise multiplication (denoted as $\otimes$) as follows:
\begin{align}
f^{(q)}_i &= \text{XLM-RoBERTa}(q_i) \\
f^{(v)}_i &= \text{ResNet}(v_i)\\
f^{(u)}_i &= f^{(v)}_i \otimes f^{(q)}_i
\end{align}
\mj{In addition, we employ a multi-layer perceptron (MLP) to transform the unified feature vector $f^{(u)}_i$ into a logit value $f^{(u^\prime)}_i$. To introduce non-linearity into the process, the ReLU function is utilized.} 
\begin{equation}
  f^{(u^\prime)}_i = W^{(\alpha)}\text{ReLU}(W^{(\beta)}f^{(u)}_i+b^{(\beta)})+b^{(\alpha)}  
\end{equation}
\mj{If we denote $c$ as a specific class, where $c \in \{0, 1, 2, ..., C\}$, the final predicted value $\hat{y}_i$ can be obtained by selecting the class with the highest probability after applying the Softmax transformation. This can be expressed as follows:}
\begin{equation}
  \hat{y}_{i} = \argmax_{c}(\text{Softmax}(f^{(u^\prime)}_{i,c}))  
\end{equation}

\subsection{Triple Prediction Module}\label{subsec:TPModule}
\begin{figure}[h!]
  \includegraphics[width=\linewidth]{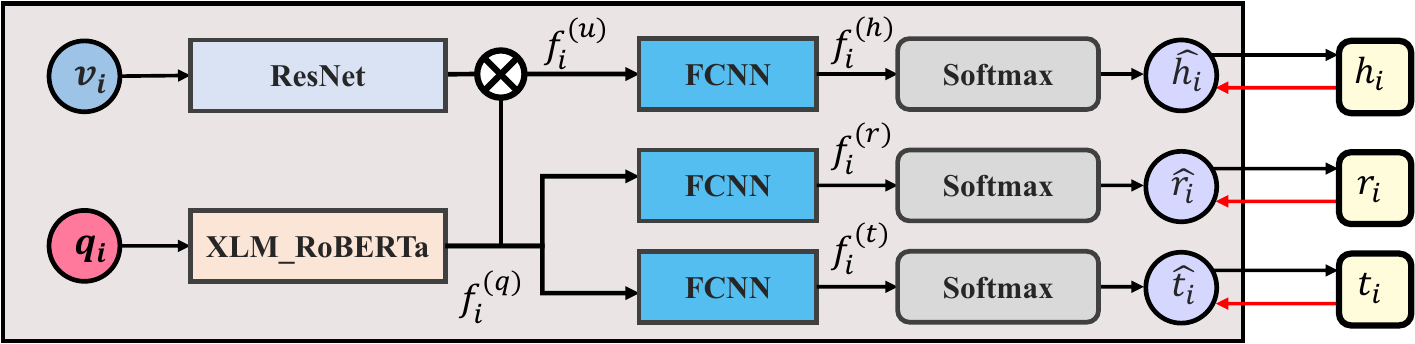}
  \caption{Architecture of Triple Prediction Module.}
  \label{fig:TP Module}
\end{figure}

Figure ~\ref{fig:TP Module} demonstrates the triple prediction module. It predicts the head, relation, and tail from the given image and question. Considering the features of each triple, the proposed model implements shared parameters for the image and question feature extractors. Concurrently, it maintains distinct parameters for each classifier responsible for predicting the head, relation, and tail. In other words, the classifier processes each head, relation, and tail prediction using a different linear layer. As mentioned in the section of Proposed BOK-VQA Dataset, the triple prediction module demonstrates language-independent characteristics. This is because the proposed method yields one-hot encoded target values regardless of their source language.

\subsection{KGE Module}\label{subsec:KGEModule}

\kt{KGE is an approach for transforming KGs from the triple to embedding form.} Hence, similar to word embeddings, 
\kt{KGE} must be pretrained before training. The KGE pretraining process can be divided into two steps, i.e., sampling and training. First, in the sampling step, assuming that the triple information included in the training data is a valid triple set $\mathcal{K}$, an invalid triple set $\mathcal{K}^\prime$ is generated by randomly replacing the head or tail. The final valid/invalid triple set can be defined as $\mathbb{K} = \{\mathcal{K} \cup \mathcal{K}^\prime\}$. Second, in the training step, the created invalid triples are trained using a loss function $(\mathcal{L}_{c})$ that effectively distinguishes between valid and invalid triples as follows :
\begin{align}
    \mathcal{L}_{c} &= \sum_{(h, r, t) \in \mathbb{K}} \log (1+e^{g(h, r, t)}) + \frac{\lambda}{2}||w||_2^2\\
    g(h, r, t) &=\begin{cases}
    		f(h, r, t),  & \text{for $(h, r, t) \in \mathcal{K}$ }\\
                -f(h, r, t), & \text{for $(h, r, t) \in \mathcal{K}^\prime$}
    	    \end{cases} \\
    f(h, r, t) &= W\cdot(\text{ReLU}([e_h; e_r; e_t] * \Omega)) + b    
\end{align}

In this study, we used the ConvKB approach, which captures the patterns of individual entities and relations present in a KG via convolution operations. The aforementioned function $f$ signifies the scoring function employed by ConvKB. Within this context, $*$ represents the convolution operation, while $e_h, e_r, e_t$ denote head, relation, and tail embeddings, respectively. They are randomly initialized in the beginning, similar to word embeddings. Furthermore, $\Omega$ represents the set of weights used in the convolutional filter. We trained ConvKB over our entire KG (280k) for 50,000 iterations.

\subsection{Aligning KG for VQA}\label{subsec:AlingnKG}
\kt{When KGEs are successfully trained using the KGs over our dataset, pretrained knowledge embeddings} corresponding to triples $h_i$, $r_i$, and $t_i$ which are necessary for the answer from the image $v_i$ and question $q_i$ can be represented as $e_{h_i}, e_{r_i},$ and $e_{t_i}$, respectively, as follows:
\begin{align}
  emb_{kb_i} = \{e_{h_i}, e_{r_i}, e_{t_i}\} &= ConvKB(h_i, r_i, t_i)\\
  f^{(z)}_i &= f^{(u)}_i\otimes emb_{kb_i}  
\end{align}

Where $f^{(u)}_i$ denotes the fusion feature vector of modalities, which is an element-wise product of image feature vector $f^{(v)}_i$ and question feature vector $f^{(q)}_i$.

\subsection{GEL-VQA}\label{subsec:GEL-VQA}

\begin{figure}[H]
\centering
\includegraphics[width=\linewidth]{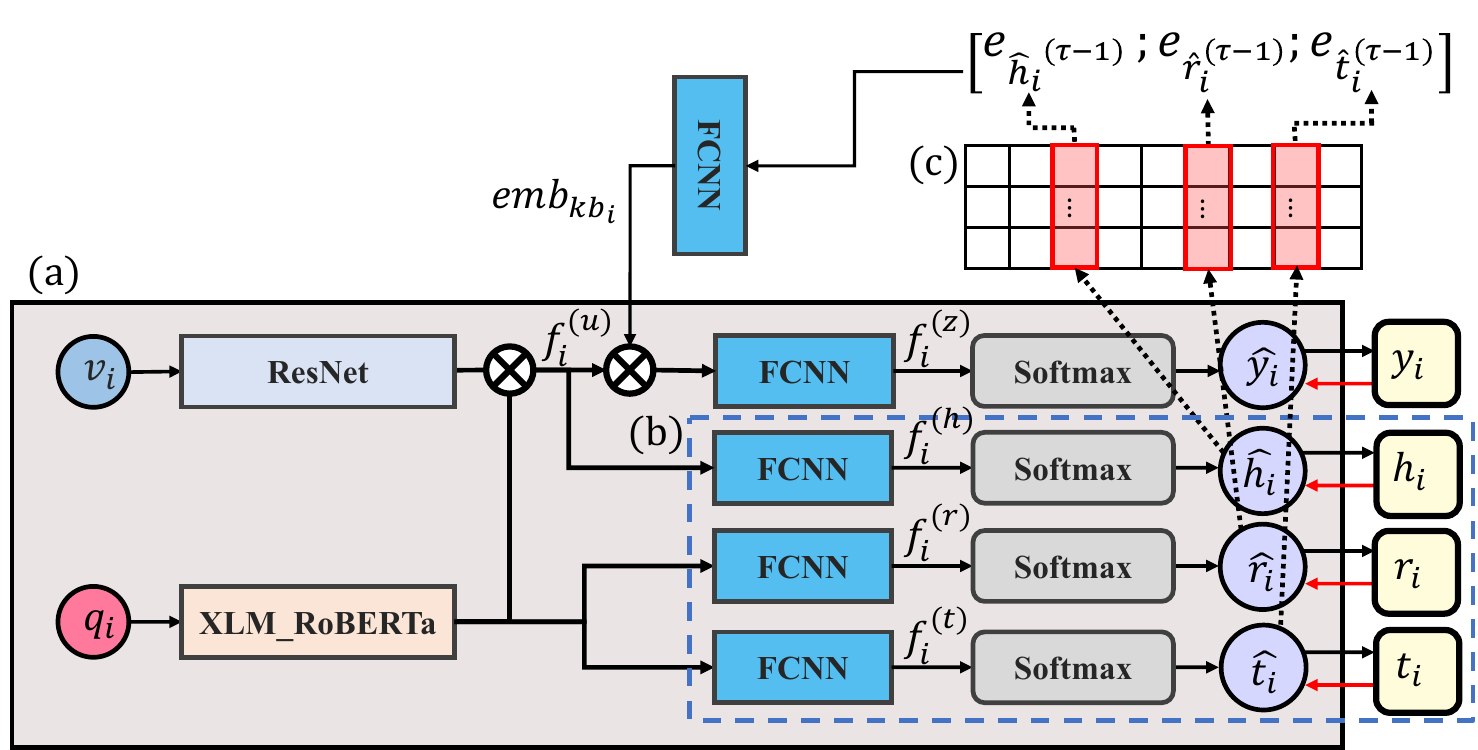}
  \caption{GEL-VQA architecture: (a) VQA module, (b) triple prediction module, and (c) pretrained KGE.}
\label{fig:GEL-VQA}
\end{figure}

\begin{table*}[h]
\centering
\small
\begin{tabular}{cccccc}
\toprule
      Language      & \textsc{baseline} & \textsc{gel-vqa(Ideal)}        & \textsc{gel-vqa}       & \textsc{gel-vqa} + TF & \textsc{gel-vqa} + TF + ATTN   \\
\midrule 
      Bilingual     &21.51$\pm$1.81     &66.01$\pm$1.83                  &45.08$\pm$0.94          &48.07$\pm1$.33         &48.11$\pm$1.50   \\ 
      English       &21.90$\pm$2.34     &66.68$\pm$1.15                  &47.83$\pm$0.56          &51.64$\pm$0.88         &50.74$\pm$0.90 \\
      Korean        &21.16$\pm$1.50     &72.25$\pm$1.29                  &50.30$\pm$2.24          &53.40$\pm$2.73         &55.48$\pm$1.89   \\  

\bottomrule
\end{tabular}
\caption{Performance of five models across three languages: Bilingual, English, and Korean.}
\label{tab:table1}
\end{table*}

In the context of VQA that uses external knowledge, it is unrealistic to assume that one possesses external knowledge pertaining to the given images and questions. Consequently, in this study, we proposed the \textsc{gel-vqa} model that employs a multitask learning approach to perform triple prediction and uses the predicted triples as external knowledge. To predict and use triples in real time, the model utilizes the predicted triples $\{\hat{h}_i^{(\tau-1)}, \hat{r}_i^{(\tau-1)}, \hat{t}_i^{(\tau-1)}\}$, computed using updated weights from the previous time step $\tau -1$.
Figure~\ref{fig:GEL-VQA} illustrates the structure of \textsc{gel-vqa}. At time step $\tau \ge 1$, the given data $v_i, q_i$ and KGE $\mathcal{E}^{(\tau-1)}_i = [e_{\hat{h}^{(\tau-1)}_i}; e_{\hat{r}^{(\tau-1)}_i}; e_{\hat{t}^{(\tau-1)}_i}]$ predicted through triple prediction at time-step $\tau-1$ can be used as external knowledge inputs. \kt{Where $\mathcal{E}^{(0)}_i$ contains random embeddings of head, relation, and tail, and `;' denotes the concatenation of its embeddings.}
\begin{multline} \label{gel-vqa-loss}
\mathcal{L}_{\mathrm{VQA}}^{(\tau)}(\theta) =  - \frac{1}{N} \sum_{i=1}^{N}\sum_{j=1}^{C}\\
y_{i, j} \log(P_{vqa}(j | v_i, q_i, \mathcal{E}^{(\tau-1)}_i; \theta))
\end{multline}

\kt{$\mathcal{L}_{\mathrm{VQA}}^{(\tau)}$ in equation ~\eqref{gel-vqa-loss} represents the loss function of the VQA module at time-step $\tau$, where image $v_i$, question $q_i$, and embedding $\mathcal{E}^{(\tau-1)}_i$ of the triple predicted in the previous time-step $\tau-1$ are used as inputs. }
\kt{Similarly, $\mathcal{L}_{\mathrm{T}}(\theta)$ in equation ~\eqref{gel-vqa-loss-total}} denotes the loss function of the triple prediction module. \kt{Finally, the total loss function of our \textsc{gel-vqa}, $\mathcal{L}^{(\tau)}(\theta)$,} can be represented as:
\begin{equation} \label{gel-vqa-loss-total}
    \mathcal{L}^{(\tau)}(\theta) = \mathcal{L}_{\mathrm{VQA}}^{(\tau)}(\theta) + \mathcal{L}_{\mathrm{T}}(\theta)
\end{equation}

\section{Experiments}
Our research objective involves exploring the feasibility of flexible multilingual VQA using KGs of high-resource languages. This can be accomplished through experiments using the proposed BOK-VQA dataset and \textsc{gel-vqa} model to address the following questions: (1) Does the proposed BOK-VQA dataset require real-world external knowledge? (2) Does the proposed \textsc{gel-vqa} model utilize KGs independent of language? (3) \kt{Is there a performance degradation in relatively less-resourced languages based on the proposed \textsc{gel-vqa}?}

\subsection{Experimental Settings}\label{subsec:experiment-setting}
We performed experiments using five models for three different language combinations. The dataset was split into 60\% for training and 20\% each for validation and testing, with five-fold validation. 
The proposed dataset contains one question-answer pair per sample; therefore, all evaluation metrics used in the experiment utilized accuracy.
First, for the Korean evaluation, we used only 17K Korean questions and an English KB for training. Second, for English, we used the English KB and 17K English questions. Conversely, in the bilingual combination, both 17K Korean and 17K English question sets were used in conjunction with the English KB for training. As introduced in KGE module, to utilize triple embedding in VQA experiments, all experiments were pretrained on English KB data using the ConvKB algorithm for 50,000 iterations. 
Detailed settings of the hyperparameters applied in the experiments are presented in Appendix C. The proposed VQA models are described below. 

\begin{itemize}
    \item \textsc{baseline}: A basic VQA model that utilizes image-question pairs without employing external knowledge.
    \item \textsc{gel-vqa(Ideal)}: Given an image-question corresponding to a triple, this model uses ConvKB to embed triples. The model operates under the assumption that it is aware of all triples in the training/testing data and accordingly infers VQA answers. Owing to this assumption, it may not be suitable for real-world applications. However, by examining the difference between \textsc{gel-vqa(Ideal)} and \textsc{baseline} this model demonstrated that the BOK-VQA dataset used in this study requires external knowledge.
    \item \textsc{gel-vqa}: This model is an evolved version of \textsc{gel-vqa(Ideal)} through triple prediction and utilization of predicted triples.
    \item \textsc{gel-vqa} + TF: This model applies the concept of “teacher forcing” to the \textsc{gel-vqa} model using ground-truth triples for training data during VQA model training. Note that the test \kt{phase} used predicted triples.
    \item \textsc{gel-vqa} + TF + ATTN:  This model strengthens selective knowledge acquisition by applying multi-head attention to the $<$head, relation, tail$>$ of KGE (see Appendix D).
\end{itemize}

\subsection{Experiment Results}
\textbf{(Overall)} Table~\ref{tab:table1} summarizes the experimental results of each model on the datasets used for training in terms of language. Overall, \textsc{baseline}, which does not utilize external knowledge, exhibits the lowest performance, whereas \textsc{gel-vqa(ideal)}, the ideal case, exhibits the highest performance. The performance gap between the \textsc{baseline} and \textsc{gel-vqa(ideal)} models suggests that the BOK-VQA dataset is more effective when external knowledge is used. Additionally, the performance of the proposed model, \textsc{gel-vqa}, is 25.97 points higher than that of \textsc{baseline}, indicating that it could effectively utilize the external knowledge provided by KGs.

\noindent \textbf{(Performance in Less-resourced Languages)} Table~\ref{tab:table1} indicates that despite the use of English KB, the performance in Korean, a relatively less-resourced language was on average 3.7\% higher than that in English. This indicates that the training performance of less-resourced languages does not deteriorate, and the quality of question construction is more important. 

\noindent \textbf{(Language-independent KG Utilization)} Table~\ref{tab:table2} summarizes the performance of the proposed triple prediction method. Despite the use of English KGs, the performance of predicting triples based on images and questions in Korean was similar to or slightly better than that in English. These results imply that KGs are independent of language. However, it should be noted that the English question data were constructed based on Korean questions, which were constructed first; thus, Korean sentences may have been more natural in terms of quality and structure. 

\begin{table}[t]
\centering
\small
\begin{tabular}{ccccccc}
\toprule
         Language      & Head            & Relation              & Tail                & Overall     \\
\midrule 
         Bilingual     &63.82            &82.76                  & 70.73               & 47.17       \\ 
         English       &66.24            &81.93                  & 68.51               & 47.59       \\
         Korean        &65.37            &85.39                  & 75.08               & 50.43       \\  

\bottomrule
\end{tabular}
\caption{Triple prediction performance of the \textsc{gel-vqa} model}
\label{tab:table2}
\end{table}
\section{Analysis}

In this section, we analyze the quality of the proposed dataset and methods of knowledge utilization used in the model.

\subsection{Robustness Test of Proposed Dataset}

\begin{table}[h]
\centering
 \small
\begin{tabular}{lcc}
    \toprule
Eval Method     & \textsc{baseline}   & \textsc{gel-vqa}   \\      
\midrule
Raw Question            &21.51$\pm$1.81     &45.08$\pm$0.94 \\     
Punctuation             &21.09$\pm$0.81     &44.02$\pm$0.93 \\     
Antonyms                &20.94$\pm$0.50      &37.23$\pm$1.62 \\     
Synonym(Verb)           &21.07$\pm$0.77     &43.36$\pm$0.90 \\     
Synonym(Noun)           &20.97$\pm$0.59     &42.79$\pm$0.91 \\     
Hypernym(Noun)          &20.76$\pm$0.23     &36.24$\pm$0.76 \\     
Hyponym(Noun)           &20.83$\pm$0.33     &38.23$\pm$1.01 \\     
    \bottomrule
\end{tabular}
\caption{Model robustness test results through question transformation.}
\label{tab:robustness}
\end{table}

High-quality data should be unbiased and robust. How, then, can we analyze the quality of the BOK-VQA dataset proposed in this study? CARETS \cite{jimenez2022carets} and VALSE \cite{parcalabescu-etal-2022-valse} are methods used to determine the consistency and robustness of a model by transforming questions into VQA. For example, a model using consistent data should consistently answer the questions “Who is Barack\_Obama’s \textbf{wife}?” and “Who is Barack\_Obama’s \textbf{spouse}?”, while data with errors or conflicts would not. To verify the consistency and robustness of our model, we conducted robustness experiments through six types of question modifications based on WordNet \cite{wordnet, nltk}, as shown in column “Eval Method” in Table~\ref{tab:robustness}. 

Question transformation involves changing a word in a question using the aforementioned six methods. For example, in the case of noun synonyms, the system sequentially searches for nouns in a question and replaces them with synonyms from WordNet, if available. Table~\ref{tab:robustness} lists the results of experiments on data robustness. When nouns are replaced with hyponyms or hypernyms, the performance significantly drops in comparison to that of the original question, which is used as the baseline. Specifically, in the case of hypernyms, replacing “wife” with “person” increases the ambiguity when searching for triples in the transformed sentence, resulting in performance deterioration. 

In contrast, when nouns are replaced with synonyms, such as changing “wife” to “spouse,” the performance remains relatively similar to that of the original question. This indicates that preserving meaning through synonym replacement does not adversely affect the system’s ability to retrieve appropriate information from KGs. Transformations, including punctuation and verb changes, demonstrated similar performance as the original question test.

In conclusion, performance deterioration due to noun replacement showed that nouns in questions have a strong correlation with objects in images or heads in KGs. When the meaning of a noun changes significantly, the KG system may struggle to find the desired information in the KG owing to changes in a semantic association. Finally, when comparing the quality with the VQA v2.0 dataset \cite{balanced_vqa_v2}, we observed a performance drop of 12.03\% in the ontological transformation data proposed by CARETS, whereas our dataset showed a 7.85\% decrease; this could indicate that our dataset is more robust.

\subsection{Triple Attention Score Statistics}

\begin{figure}[h]
    \includegraphics[width=\linewidth]{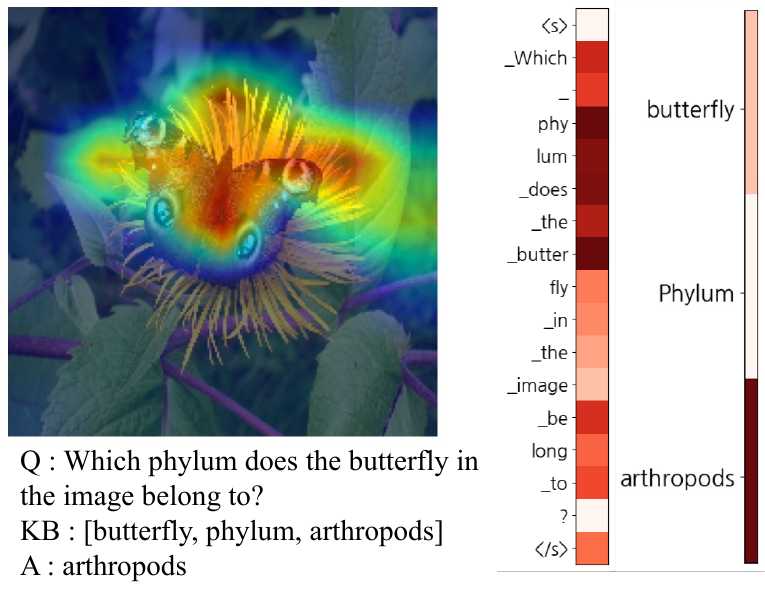}
    \caption{Visualizing attention score of H-Given case.}
    \label{fig:fig5}
\end{figure}

\begin{table}[h]
\centering
\small

\begin{tabular}{cccc}
\toprule
    Case              & H attention   & R attention  & T attention   \\
\midrule 
    H-Given           & 0.3239        & 0.2587       &0.4174       \\
    T-Given           & 0.3280        & 0.2790       &0.3930         \\
    HT-Given          & 0.3329        & 0.2819       &0.3852         \\
\bottomrule
\end{tabular}
\caption{Average attention scores based on the triple component.}
\label{tab:attn-score}
\end{table}

In the proposed \textsc{gel-vqa}+ATTN model, self-attention \cite{vaswani2017attention} was applied to the head, relation and tail as mentioned in the section of Experiment Settings.
Consequently, when solving the VQA problem, we could examine the information (between ${e_{h_i}, e_{r_i}, e_{t_i}}$) that the model focused on more. Figure~\ref{fig:fig5} shows which information between $<$head, relation, tail$>$ the model focused on for a single sample. Including this example, in most cases, the attention scores were the highest for the tail. This is because the `tail' often represents the objective or outcome of the ``relation''; hence, the model primarily sought outcome information through external knowledge. Interestingly, when the question included information from $<$h, r, t$>$, the model tended to focus slightly more on triple components not included in the question. For instance, Figure~\ref{fig:fig5} shows a question that includes head information; in this case, the model tended to focus relatively more on information other than the head.

We analyzed the results presented in Table~\ref{tab:attn-score} based on three criteria (H-, T-, and HT-Given) to verify the extent of knowledge utilization in the model. Table~\ref{tab:attn-score} lists the average attention scores for the entire dataset under three cases.

\begin{itemize}
\item \textbf{H-Given}: This refers to cases in which the head information was included in a question. The attention score was the highest for the tail, which suggests that because the model had already obtained information about the head from the question, it focused more on obtaining tail information from external knowledge.
\item \textbf{T-Given}: This refers to cases in which the tail information was included in the question. The attention scores for both head and tail were relatively evenly distributed. This suggests that because the model had already obtained information about the tail from the question, it tended to focus on the head and relation in the external knowledge.

\item \textbf{HT-Given}: This refers to cases in which both head and tail information were included in the question. In this case, the model tended to focus more on the part corresponding to “relation” in the embeddings given by external knowledge. Therefore, compared to H-Given and T-Given, the attention score for “relation” appeared to be higher.
\end{itemize}

In conclusion, although the differences in attention scores were marginal, the triple information on which the model focuses varied depending on the type of question.

\subsection{Analysis of KGE Model Impact on GEL-VQA Performance}

The foremost contribution of this study is the construction of a language-independent KB-VQA using KGEs. Various KGE methods have been proposed for this purpose, which raises the question: what impact does the choice of different KGE training methods have on VQA, and which KGE is the most efficient? Previous studies utilized FB15K \cite{rossi2021knowledge}, a large relational graph dataset, to evaluate the performance of different KGE training methods. FB15K employs link and entity prediction methods for the intrinsic evaluation of KGE models using Hit@10 as an evaluation metric that determines whether the correct answer is included in the top 10 predicted results. The FB15K column in Table~\ref{tab:table6} presents the performance evaluation of various KGE training methods using FB15K, specifically the Hit@10 score. In this study, the performance of ConvKB, used as a baseline KB, was significantly different (more than two-fold) in comparison to that of recently proposed algorithms under the FB15K dataset. Conversely, \textsc{gel-vqa} (Table~\ref{tab:table6}), which represents the intrinsic evaluation of corresponding KGEs in VQA, exhibited a relatively lower difference in performance between ConvKB and other models. The Pearson correlation coefficient between the scores of FB15K(H@10) and proposed \textsc{gel-vqa-Ideal} model had a high value of 0.85. This implies a correlation between the performance of the two benchmarks, signifying that the choice of the KGE model significantly affects the performance of both \textsc{gel-vqa} and \textsc{gel-vqa-Ideal}. These results reveal the potential applicability of KGs and indicate the need for further research to effectively harness KGEs. 
\begin{table}[t!]
\centering
\small
\begin{tabular}{ccccccc}
\toprule
         KGE   & \textsc{gel-vqa}           & \textsc{gel-vqa(Ideal)}         & FB15K 
         \\        
\midrule 
         TransE       &51.22$\pm$1.02              &79.56$\pm$0.77                      & 0.847                \\
         TorusE       &50.51$\pm$1.32              &79.15$\pm$0.28                      & 0.839                \\
         HolE         &48.64$\pm$0.75              &78.91$\pm$1.08                      & 0.867                \\
         DistMul      &47.28$\pm$0.80              &72.40$\pm$1.60                      & 0.863                \\
         ConvKB       &45.08$\pm$0.94              &66.01$\pm$1.83                      & 0.408                \\

\bottomrule
\end{tabular}
\caption{Impact of KGE methods on \textsc{gel-vqa}.}
\label{tab:table6}
\end{table}
\section{Conclusion}

In this study, we proposed a method to effectively construct external knowledge VQA training data for less-resourced languages using the abundant external knowledge of high-resource languages. We constructed large-scale training data using 17K Korean-English question-answer pairs and 280K instances of information. Furthermore, we demonstrated a language-independent approach to KG utilization using a \textsc{gel-vqa} model that performed VQA and KGE training in a multitasking manner. Our experiments demonstrated the acceptable performance of a bilingual VQA model using the proposed BOK-VQA and \textsc{gel-vqa}. Through the results of this study, 
we anticipate that the benefits of multilingual training and our evaluation approach can serve as performance metrics for future multilingual VQA research. Nonetheless, there are several unresolved limitations in this study. While the knowledge utilized to construct a question involves K $\ge$ 1 pieces of information, our study predicts and utilizes only one piece of knowledge. Thus, it remains inadequate for solving problems that require the complex utilization of multiple pieces of knowledge. 

\section*{Acknowledgements}
This research was supported by the National Research Foundation of Korea (2021R1F1A1063474) for KyungTae Lim. This research used datasets from The Open AI Dataset Project (AI-Hub) (No. 2022-데이터-위41, 2023-지능데이터-위93).
\bibliography{aaai24}

\begin{thebibliography}{32}
\providecommand{\natexlab}[1]{#1}

\bibitem[{Antol et~al.(2015)Antol, Agrawal, Lu, Mitchell, Batra, Zitnick, and Parikh}]{antol2015vqa}
Antol, S.; Agrawal, A.; Lu, J.; Mitchell, M.; Batra, D.; Zitnick, C.~L.; and Parikh, D. 2015.
\newblock Vqa: Visual question answering.
\newblock In \emph{Proceedings of the IEEE international conference on computer vision}, 2425--2433.

\bibitem[{Auer et~al.(2007)Auer, Bizer, Kobilarov, Lehmann, Cyganiak, and Ives}]{auer2007dbpedia}
Auer, S.; Bizer, C.; Kobilarov, G.; Lehmann, J.; Cyganiak, R.; and Ives, Z. 2007.
\newblock Dbpedia: A nucleus for a web of open data.
\newblock In \emph{The Semantic Web: 6th International Semantic Web Conference, 2nd Asian Semantic Web Conference, ISWC 2007+ ASWC 2007, Busan, Korea, November 11-15, 2007. Proceedings}, 722--735. Springer.

\bibitem[{Bordes et~al.(2013)Bordes, Usunier, Garcia-Duran, Weston, and Yakhnenko}]{transE}
Bordes, A.; Usunier, N.; Garcia-Duran, A.; Weston, J.; and Yakhnenko, O. 2013.
\newblock Translating embeddings for modeling multi-relational data.
\newblock \emph{Advances in neural information processing systems}, 26.

\bibitem[{Choi et~al.(2021)Choi, On, Heo, Seo, Jang, Lee, and Zhang}]{choi2021dramaqa}
Choi, S.; On, K.-W.; Heo, Y.-J.; Seo, A.; Jang, Y.; Lee, M.; and Zhang, B.-T. 2021.
\newblock DramaQA: Character-centered video story understanding with hierarchical qa.
\newblock In \emph{Proceedings of the AAAI Conference on Artificial Intelligence}, volume~35, 1166--1174.

\bibitem[{Conneau et~al.(2020)Conneau, Khandelwal, Goyal, Chaudhary, Wenzek, Guzm{\'a}n, Grave, Ott, Zettlemoyer, and Stoyanov}]{xlmroberta}
Conneau, A.; Khandelwal, K.; Goyal, N.; Chaudhary, V.; Wenzek, G.; Guzm{\'a}n, F.; Grave, {\'E}.; Ott, M.; Zettlemoyer, L.; and Stoyanov, V. 2020.
\newblock Unsupervised Cross-lingual Representation Learning at Scale.
\newblock In \emph{Proceedings of the 58th Annual Meeting of the Association for Computational Linguistics}, 8440--8451.

\bibitem[{Deng et~al.(2009)Deng, Dong, Socher, Li, Li, and Fei-Fei}]{imagenet}
Deng, J.; Dong, W.; Socher, R.; Li, L.-J.; Li, K.; and Fei-Fei, L. 2009.
\newblock ImageNet: A large-scale hierarchical image database.
\newblock In \emph{2009 IEEE Conference on Computer Vision and Pattern Recognition}, 248--255.

\bibitem[{Dettmers et~al.(2018)Dettmers, Minervini, Stenetorp, and Riedel}]{ConvE}
Dettmers, T.; Minervini, P.; Stenetorp, P.; and Riedel, S. 2018.
\newblock Convolutional 2d knowledge graph embeddings.
\newblock In \emph{Proceedings of the AAAI conference on artificial intelligence}, volume~32.

\bibitem[{Goyal et~al.(2017)Goyal, Khot, Summers{-}Stay, Batra, and Parikh}]{balanced_vqa_v2}
Goyal, Y.; Khot, T.; Summers{-}Stay, D.; Batra, D.; and Parikh, D. 2017.
\newblock Making the {V} in {VQA} Matter: Elevating the Role of Image Understanding in {V}isual {Q}uestion {A}nswering.
\newblock In \emph{Conference on Computer Vision and Pattern Recognition (CVPR)}.

\bibitem[{He et~al.(2016)He, Zhang, Ren, and Sun}]{resnet}
He, K.; Zhang, X.; Ren, S.; and Sun, J. 2016.
\newblock Deep residual learning for image recognition.
\newblock In \emph{Proceedings of the IEEE conference on computer vision and pattern recognition}, 770--778.

\bibitem[{Hudson and Manning(2019)}]{hudson2019gqa}
Hudson, D.~A.; and Manning, C.~D. 2019.
\newblock Gqa: A new dataset for real-world visual reasoning and compositional question answering.
\newblock In \emph{Proceedings of the IEEE/CVF conference on computer vision and pattern recognition}, 6700--6709.

\bibitem[{Jimenez, Russakovsky, and Narasimhan(2022)}]{jimenez2022carets}
Jimenez, C.; Russakovsky, O.; and Narasimhan, K. 2022.
\newblock CARETS: A Consistency And Robustness Evaluative Test Suite for VQA.
\newblock In \emph{Proceedings of the 60th Annual Meeting of the Association for Computational Linguistics (Volume 1: Long Papers)}, 6392--6405.

\bibitem[{Krishna et~al.(2017)Krishna, Zhu, Groth, Johnson, Hata, Kravitz, Chen, Kalantidis, Li, Shamma et~al.}]{visualgenome}
Krishna, R.; Zhu, Y.; Groth, O.; Johnson, J.; Hata, K.; Kravitz, J.; Chen, S.; Kalantidis, Y.; Li, L.-J.; Shamma, D.~A.; et~al. 2017.
\newblock Visual genome: Connecting language and vision using crowdsourced dense image annotations.
\newblock \emph{International journal of computer vision}, 123: 32--73.

\bibitem[{Lehmann et~al.(2015)Lehmann, Isele, Jakob, Jentzsch, Kontokostas, Mendes, Hellmann, Morsey, Van~Kleef, Auer et~al.}]{lehmann2015dbpedia}
Lehmann, J.; Isele, R.; Jakob, M.; Jentzsch, A.; Kontokostas, D.; Mendes, P.~N.; Hellmann, S.; Morsey, M.; Van~Kleef, P.; Auer, S.; et~al. 2015.
\newblock Dbpedia--a large-scale, multilingual knowledge base extracted from wikipedia.
\newblock \emph{Semantic web}, 6(2): 167--195.

\bibitem[{Lei et~al.(2018)Lei, Yu, Bansal, and Berg}]{lei2018tvqa}
Lei, J.; Yu, L.; Bansal, M.; and Berg, T. 2018.
\newblock TVQA: Localized, Compositional Video Question Answering.
\newblock In \emph{Proceedings of the 2018 Conference on Empirical Methods in Natural Language Processing}, 1369--1379.

\bibitem[{Loper and Bird(2002)}]{nltk}
Loper, E.; and Bird, S. 2002.
\newblock NLTK: the Natural Language Toolkit.
\newblock In \emph{Proceedings of the ACL-02 Workshop on Effective tools and methodologies for teaching natural language processing and computational linguistics-Volume 1}, 63--70.

\bibitem[{Miller(1995)}]{wordnet}
Miller, G.~A. 1995.
\newblock WordNet: a lexical database for English.
\newblock \emph{Communications of the ACM}, 38(11): 39--41.

\bibitem[{Nguyen et~al.(2018)Nguyen, Nguyen, Phung et~al.}]{nguyen2018novel}
Nguyen, T.~D.; Nguyen, D.~Q.; Phung, D.; et~al. 2018.
\newblock A Novel Embedding Model for Knowledge Base Completion Based on Convolutional Neural Network.
\newblock In \emph{Proceedings of the 2018 Conference of the North American Chapter of the Association for Computational Linguistics: Human Language Technologies, Volume 2 (Short Papers)}, 327--333.

\bibitem[{Nickel et~al.(2011)Nickel, Tresp, Kriegel et~al.}]{rescal}
Nickel, M.; Tresp, V.; Kriegel, H.-P.; et~al. 2011.
\newblock A three-way model for collective learning on multi-relational data.
\newblock In \emph{Icml}, volume~11, 3104482--3104584.

\bibitem[{Parcalabescu et~al.(2022)Parcalabescu, Cafagna, Muradjan, Frank, Calixto, and Gatt}]{parcalabescu-etal-2022-valse}
Parcalabescu, L.; Cafagna, M.; Muradjan, L.; Frank, A.; Calixto, I.; and Gatt, A. 2022.
\newblock {VALSE}: A Task-Independent Benchmark for Vision and Language Models Centered on Linguistic Phenomena.
\newblock In \emph{Proceedings of the 60th Annual Meeting of the Association for Computational Linguistics (Volume 1: Long Papers)}, 8253--8280. Dublin, Ireland: Association for Computational Linguistics.

\bibitem[{Pfeiffer et~al.(2022)Pfeiffer, Geigle, Kamath, Steitz, Roth, Vuli{\'c}, and Gurevych}]{pfeiffer2022xgqa}
Pfeiffer, J.; Geigle, G.; Kamath, A.; Steitz, J.-M.; Roth, S.; Vuli{\'c}, I.; and Gurevych, I. 2022.
\newblock xGQA: Cross-Lingual Visual Question Answering.
\newblock In \emph{Findings of the Association for Computational Linguistics: ACL 2022}, 2497--2511.

\bibitem[{Rossi et~al.(2021)Rossi, Barbosa, Firmani, Matinata, and Merialdo}]{rossi2021knowledge}
Rossi, A.; Barbosa, D.; Firmani, D.; Matinata, A.; and Merialdo, P. 2021.
\newblock Knowledge graph embedding for link prediction: A comparative analysis.
\newblock \emph{ACM Transactions on Knowledge Discovery from Data (TKDD)}, 15(2): 1--49.

\bibitem[{Schwenk et~al.(2022)Schwenk, Khandelwal, Clark, Marino, and Mottaghi}]{schwenk2022okvqa}
Schwenk, D.; Khandelwal, A.; Clark, C.; Marino, K.; and Mottaghi, R. 2022.
\newblock A-okvqa: A benchmark for visual question answering using world knowledge.
\newblock In \emph{Computer Vision--ECCV 2022: 17th European Conference, Tel Aviv, Israel, October 23--27, 2022, Proceedings, Part VIII}, 146--162. Springer.

\bibitem[{Shah et~al.(2019)Shah, Mishra, Yadati, and Talukdar}]{K-VQA}
Shah, S.; Mishra, A.; Yadati, N.; and Talukdar, P.~P. 2019.
\newblock Kvqa: Knowledge-aware visual question answering.
\newblock In \emph{Proceedings of the AAAI conference on artificial intelligence}, volume~33, 8876--8884.

\bibitem[{Speer, Chin, and Havasi(2017)}]{speer2017conceptnet}
Speer, R.; Chin, J.; and Havasi, C. 2017.
\newblock Conceptnet 5.5: An open multilingual graph of general knowledge.
\newblock In \emph{Proceedings of the AAAI conference on artificial intelligence}, volume~31.

\bibitem[{Sun et~al.(2019)Sun, Deng, Nie, and Tang}]{rotatE}
Sun, Z.; Deng, Z.-H.; Nie, J.-Y.; and Tang, J. 2019.
\newblock Rotate: Knowledge graph embedding by relational rotation in complex space.
\newblock \emph{arXiv preprint arXiv:1902.10197}.

\bibitem[{Tapaswi et~al.(2016)Tapaswi, Zhu, Stiefelhagen, Torralba, Urtasun, and Fidler}]{tapaswi2016movieqa}
Tapaswi, M.; Zhu, Y.; Stiefelhagen, R.; Torralba, A.; Urtasun, R.; and Fidler, S. 2016.
\newblock Movieqa: Understanding stories in movies through question-answering.
\newblock In \emph{Proceedings of the IEEE conference on computer vision and pattern recognition}, 4631--4640.

\bibitem[{Trouillon et~al.(2016)Trouillon, Welbl, Riedel, Gaussier, and Bouchard}]{complEX}
Trouillon, T.; Welbl, J.; Riedel, S.; Gaussier, {\'E}.; and Bouchard, G. 2016.
\newblock Complex embeddings for simple link prediction.
\newblock In \emph{International conference on machine learning}, 2071--2080. PMLR.

\bibitem[{Vaswani et~al.(2017)Vaswani, Shazeer, Parmar, Uszkoreit, Jones, Gomez, Kaiser, and Polosukhin}]{vaswani2017attention}
Vaswani, A.; Shazeer, N.; Parmar, N.; Uszkoreit, J.; Jones, L.; Gomez, A.~N.; Kaiser, {\L}.; and Polosukhin, I. 2017.
\newblock Attention is all you need.
\newblock \emph{Advances in neural information processing systems}, 30.

\bibitem[{Wang et~al.(2017{\natexlab{a}})Wang, Wu, Shen, Dick, and van~den Hengel}]{KB-VQA}
Wang, P.; Wu, Q.; Shen, C.; Dick, A.; and van~den Hengel, A. 2017{\natexlab{a}}.
\newblock Explicit Knowledge-based Reasoning for Visual Question Answering.
\newblock In \emph{Proceedings of the Twenty-Sixth International Joint Conference on Artificial Intelligence}. International Joint Conferences on Artificial Intelligence Organization.

\bibitem[{Wang et~al.(2017{\natexlab{b}})Wang, Wu, Shen, Dick, and Van Den~Hengel}]{wang2017fvqa}
Wang, P.; Wu, Q.; Shen, C.; Dick, A.; and Van Den~Hengel, A. 2017{\natexlab{b}}.
\newblock Fvqa: Fact-based visual question answering.
\newblock \emph{IEEE transactions on pattern analysis and machine intelligence}, 40(10): 2413--2427.

\bibitem[{Wang et~al.(2014)Wang, Zhang, Feng, and Chen}]{kge}
Wang, Z.; Zhang, J.; Feng, J.; and Chen, Z. 2014.
\newblock Knowledge graph embedding by translating on hyperplanes.
\newblock In \emph{Proceedings of the AAAI conference on artificial intelligence}, volume~28.

\bibitem[{Yu et~al.(2019)Yu, Xu, Yu, Yu, Zhao, Zhuang, and Tao}]{yu2019activitynet}
Yu, Z.; Xu, D.; Yu, J.; Yu, T.; Zhao, Z.; Zhuang, Y.; and Tao, D. 2019.
\newblock Activitynet-qa: A dataset for understanding complex web videos via question answering.
\newblock In \emph{Proceedings of the AAAI Conference on Artificial Intelligence}, volume~33, 9127--9134.

\end{thebibliography}
\newpage

\appendix
\section{Appendix}

\begin{table*}[h]
\centering
\begin{tabular}{l@{\hspace{0.2cm}}l@{\hspace{0.2cm}}c@{\hspace{0.2cm}}l@{\hspace{0.2cm}}c}
\toprule
KB  & Relationship                              & \# of KBs                   & Examples      \\
\midrule 
\multirow{21}{*}{DBpedia}
    & \texttt{Subject}                          & 9242                        & (Fire\_extinguisher, Subject, English\_inventions)    \\ 
    & \texttt{Class (biology)}                  & 1427                        & (Puma, Class, Mammalia)                               \\
    & \texttt{Order (biology)}                  & 1393                        & (Common\_gull, Order, Charadriiformes)                \\  
    & \texttt{Phylum (biology)}                 & 1073                        & (Rattlesnake, Phylum, Chordata)                       \\ 
    & \texttt{Kingdom (biology)}                & 718                         & (Black swan, Kingdom, Animalia)                       \\
    & \texttt{Family (biology)}                 & 429                         & (Mangosteen, Family, Clusiaceae)                      \\ 
    & \texttt{Suborder (biology)}               & 254                         & (Japanese\_sea\_lion, Suborder, Caniformia)           \\  
    & \texttt{Unclassified Class (biology)}     & 231                         & (Maize, Unclassified\_class, Monocots)                \\
    & \texttt{Superorder (biology)}              & 182                         & (Clownfish, Superorder, Acanthopterygii)               \\ 
    & \texttt{Unclassified Phylum (biology)}    & 168                         & (Grape,Unclassified phylum, Angiospermae)             \\ 
    & \texttt{Subclassis (biology)}             & 143                         & (Termite, subclassis, Pterygota)                      \\ 
    & \texttt{Superfamilia (biology)}           & 123                         & (Human, Superfamilia, Hominoidea)                     \\ 
    & \texttt{Subphylum  (biology)}             & 122                         & (Shrimp, Subphylum, Crustacea)                        \\ 
    & \texttt{Unclassified  Order (biology)}    & 110                         & (Rosemary, Unclassified Order, Asterids)              \\ 
    & \texttt{Infraorder (biology)}              & 106                         & (Butterfly, Infraorder, Heteroneura)                   \\ 
    & \texttt{AtLocation}                       & 69                          & (Shark, Locate, Ocean)                                \\
    & \texttt{Subfamily (biology)}              & 61                          & (Common groundsel, Subfamily, Asteroideae)            \\
    & \texttt{Genus (biology)}                  & 58                          & (Peach, Genus, Prunus)                                \\
    & \texttt{Range (instrument)}               & 58                          & (Guitar, Range, 130)                                  \\
    & \texttt{Personnel}                        & 43                          & (Soccer, Personnel, 11)                               \\
    & \texttt{Others}                           & 224                         & -                                                     \\
\midrule
\multirow{11}{*}{ConceptNet}
    & \texttt{Subject}                          & 437                        & (Judge, Subject, Job)                                  \\ 
    & \texttt{RelatedTo}                        & 366                        & (Palace, RelatedTo, King)                              \\ 
    & \texttt{UsedFor}                          & 222                        & (Well, UsedFor, Storage of water)                      \\ 
    & \texttt{IsA}                              & 188                        & (Photographer, IsA, Artist)                            \\ 
    & \texttt{Phylum}                           & 72                         & (Sea\_anemone, Phylum, Cnidaria)                       \\ 
    & \texttt{AtLocation}                       & 56                         & (Sink, AtLocation, Kitchen)                            \\ 
    & \texttt{Class}                            & 47                         & (Pacific saury, Class, Actinopterygii)                 \\ 
    & \texttt{Order}                            & 25                         & (Bamboo, Order, Poales)                                \\ 
    & \texttt{PartOf}                           & 25                         & (Reel, PartOf, Fishing\_rod)                           \\ 
    & \texttt{CapableOf}                        & 23                         & (Judge, CapableOf, Impose\_a\_fine)                    \\ 
    & \texttt{Others}                           & 141                        & -                                                      \\ 
    
\bottomrule
\end{tabular}
\caption{BOK-VQA dataset statistics based on relationship}
\label{tab:appendix-relation-statistics}

\end{table*}

\begin{figure}[H]
    \includegraphics[width=\linewidth]{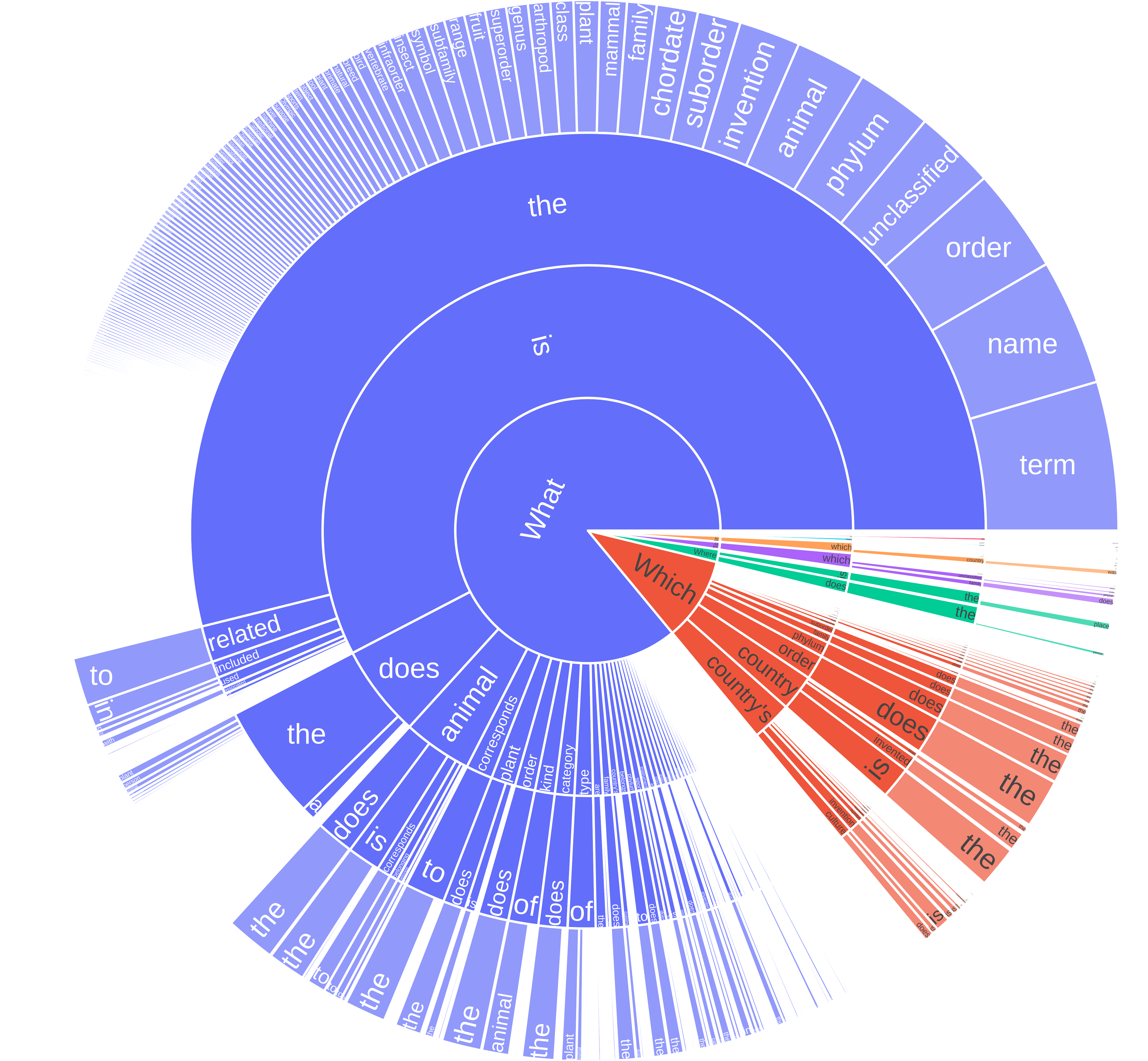}
    \captionsetup{format=plain}
    \caption{
    This is a chart illustrating the frequency of words included in the question, arranged according to their syntactical order.
  }
    \label{fig:pieplot1}
\end{figure}
\begin{figure}[H]
    \includegraphics[width=\linewidth]{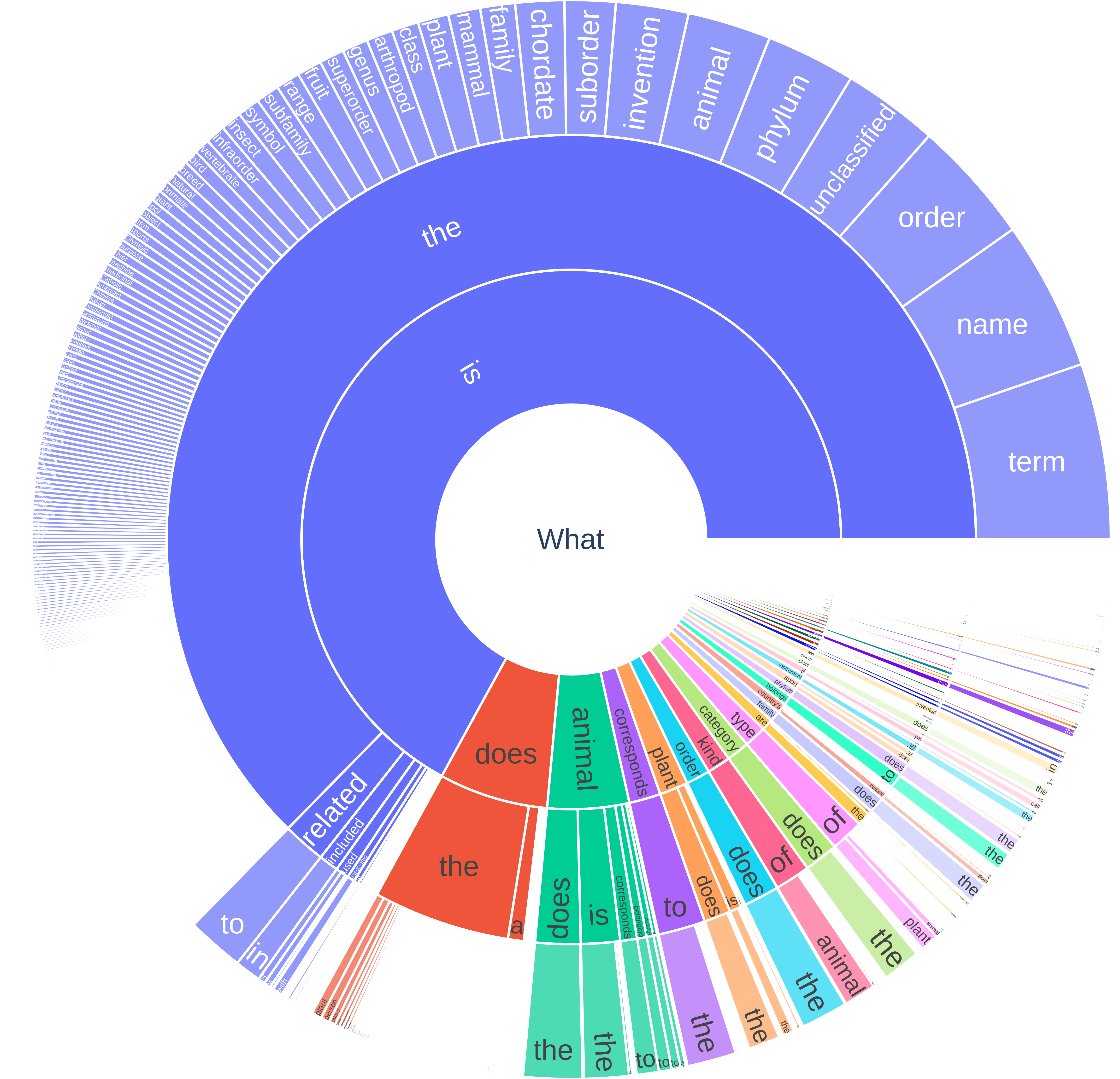}
    \captionsetup{format=plain}
    \caption{
    This chart arranges the frequency of words contained in questions starting with 'what' according to their syntactic order.
  }
    \label{fig:pieplot2}
\end{figure}

\begin{figure*}[h]
  \centering
  \begin{subfigure}{0.24\textwidth}
    \includegraphics[width=\linewidth, height=\linewidth]{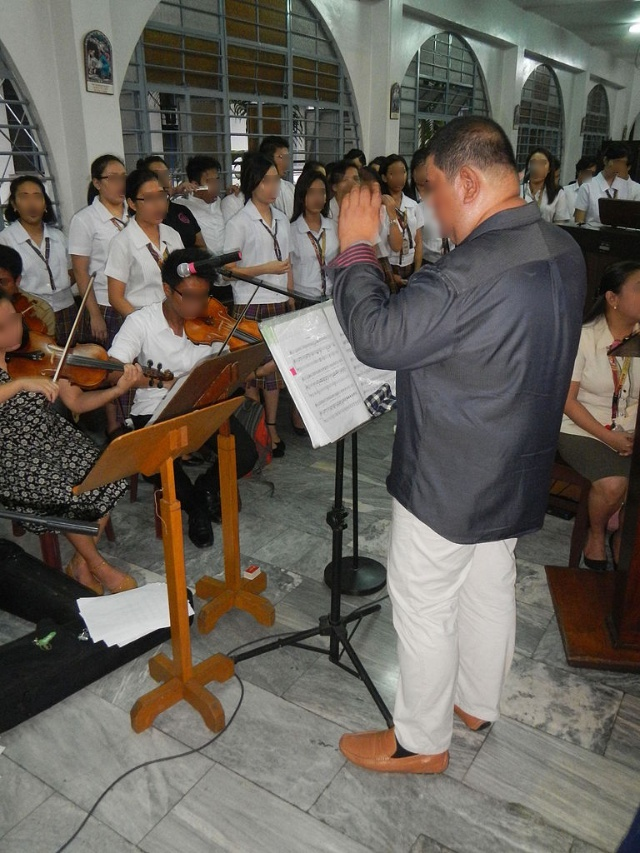}
    \scriptsize 
    \parbox{0.95\linewidth}{
    \texttt{Q : What is the range of the chordophone in the image?} \\
    \texttt{KB : [violin, range, 130]}\\
    \texttt{A : 130}\\
    \texttt{ }\\
    \texttt{ }
    }
  \end{subfigure}
  \begin{subfigure}{0.24\textwidth}
    \includegraphics[width=\linewidth, height=\linewidth]{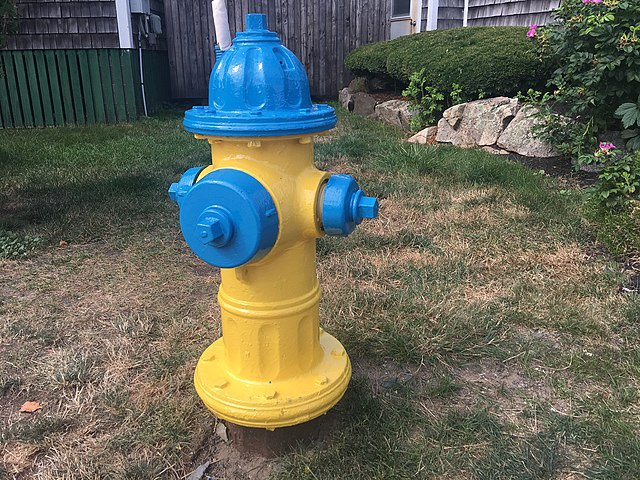}
    \scriptsize 
    \parbox{0.95\linewidth}{
    \texttt{Q : What's the firefighting equipment in the image?}\\
    \texttt{KB : [fire\_hydrant, Subject, firefighting\_equipment]}\\
    \texttt{A : fire\_hydrant}\\
    \texttt{ }
    }
  \end{subfigure}
  \begin{subfigure}{0.24\textwidth}
    \includegraphics[width=\linewidth, height=\linewidth]{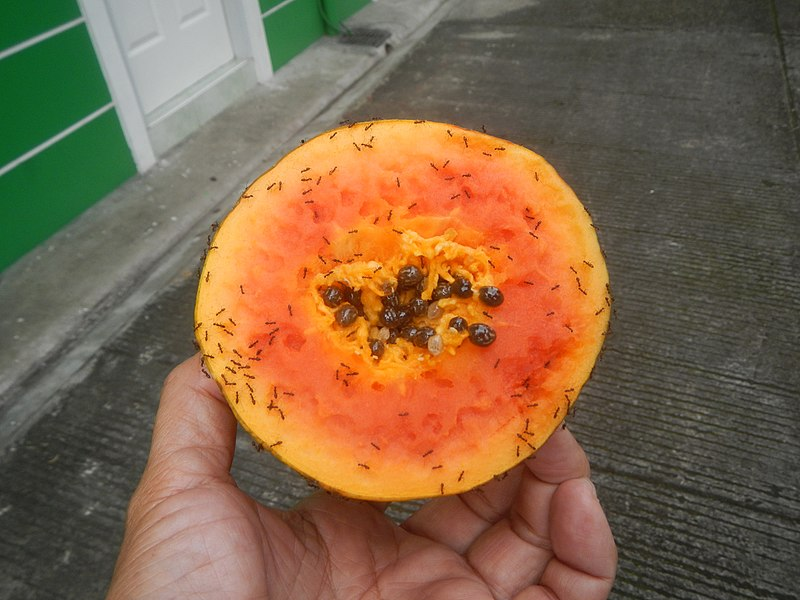}
    \scriptsize 
    \parbox{0.95\linewidth}{
    \texttt{Q : Which phylum does the animal in the image belong to?}\\
    \texttt{KB : [ant, phylum, arthropods]}\\
    \texttt{A : arthropods}\\
    \texttt{ }\\
    \texttt{ }
    }
  \end{subfigure}
  \begin{subfigure}{0.24\textwidth}
    \includegraphics[width=\linewidth, height=\linewidth]{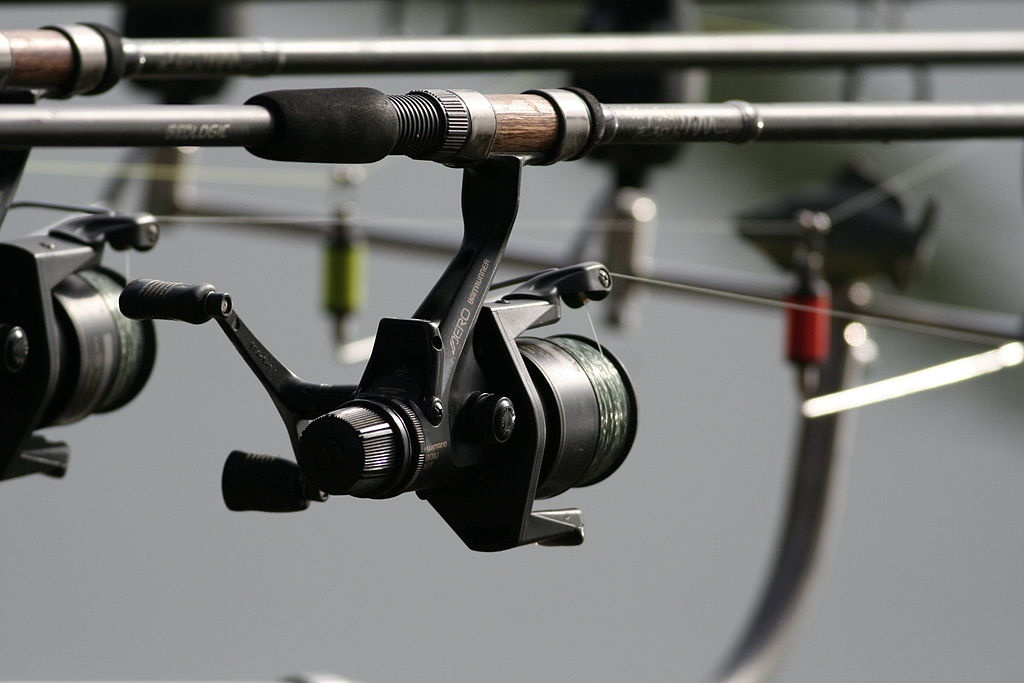}
    \scriptsize 
    \parbox{0.95\linewidth}{
    \texttt{Q : What in the image is part of a fishing rod?}\\
    \texttt{KB : [reel, PartOf, fishing\_rod]}\\
    \texttt{A : reel}\\
    \texttt{ }\\
    \texttt{ }
    }
  \end{subfigure}
  
  \begin{subfigure}{0.24\textwidth}
    \includegraphics[width=\linewidth, height=\linewidth]{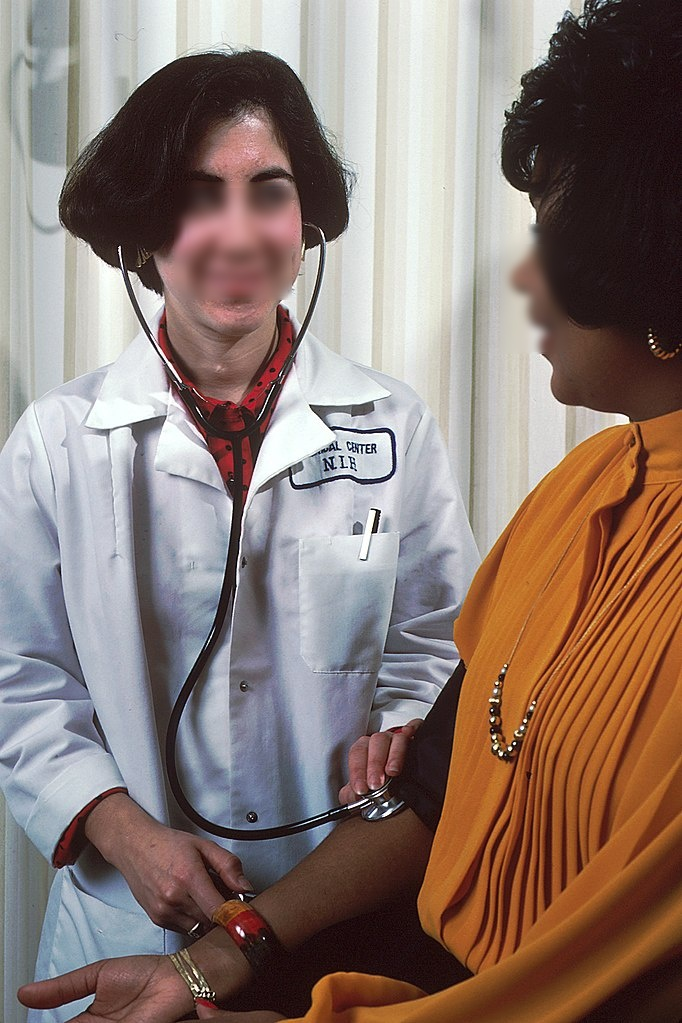}
    \scriptsize 
    \parbox{0.95\linewidth}{
    \texttt{Q : What do you call a person who can perform surgery in the image?}\\
    \texttt{KB : [doctor, CapableOf, surgery]}\\
    \texttt{A : doctor}\\
    }
  \end{subfigure}
  \begin{subfigure}{0.24\textwidth}
    \includegraphics[width=\linewidth, height=\linewidth]{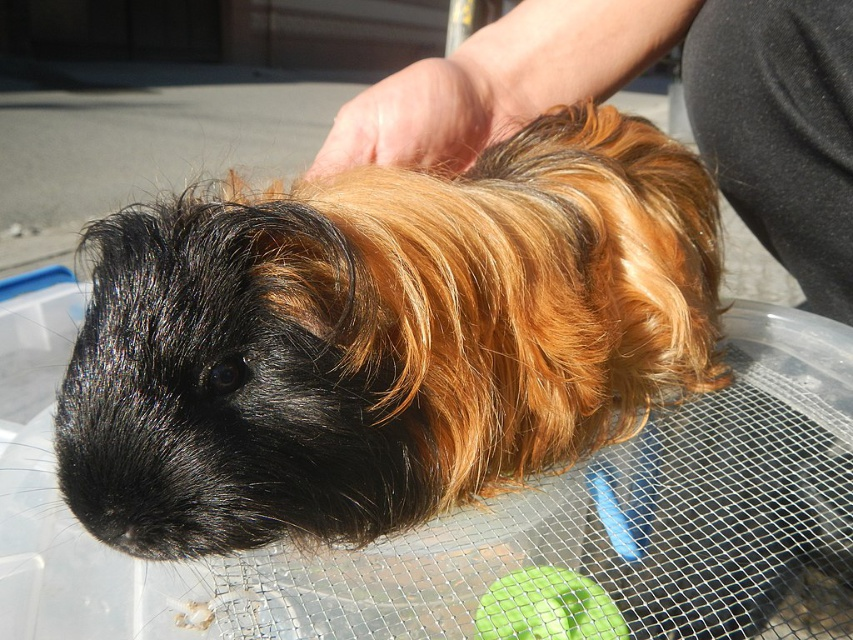}
    \scriptsize 
    \parbox{0.95\linewidth}{
    \texttt{Q : Which suborder does the pet in the image belong to?}\\
    \texttt{KB : [guinea pig, Suborder, hystricomorpha]}\\
    \texttt{A : hystricomorpha}
    }
  \end{subfigure}
  \begin{subfigure}{0.24\textwidth}
    \includegraphics[width=\linewidth, height=\linewidth]{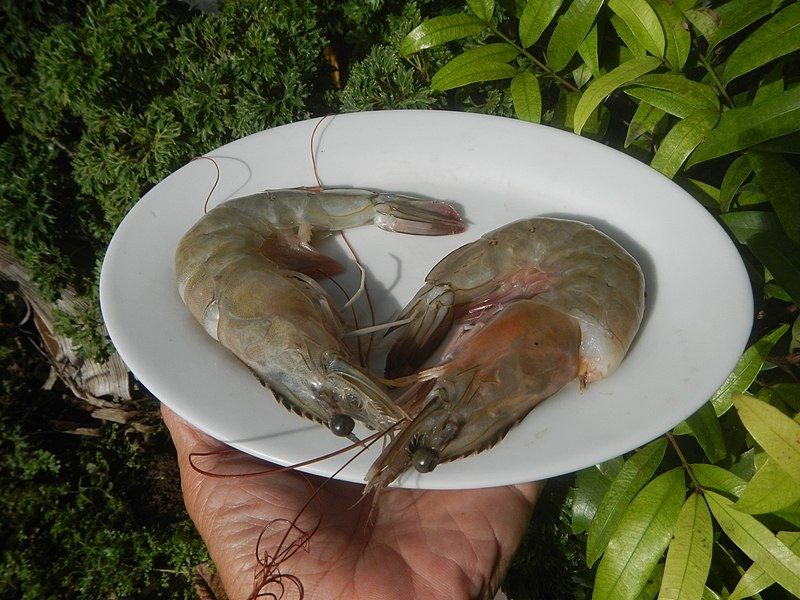}
    \scriptsize 
    \parbox{0.95\linewidth}{
    \texttt{Q : To which superorder does the animal in the image belong?}\\
    \texttt{KB : [shrimp, Superorder, eucarida]}\\
    \texttt{A : eucarida}
    
    }
  \end{subfigure}
  \begin{subfigure}{0.24\textwidth}
    \includegraphics[width=\linewidth, height=\linewidth]{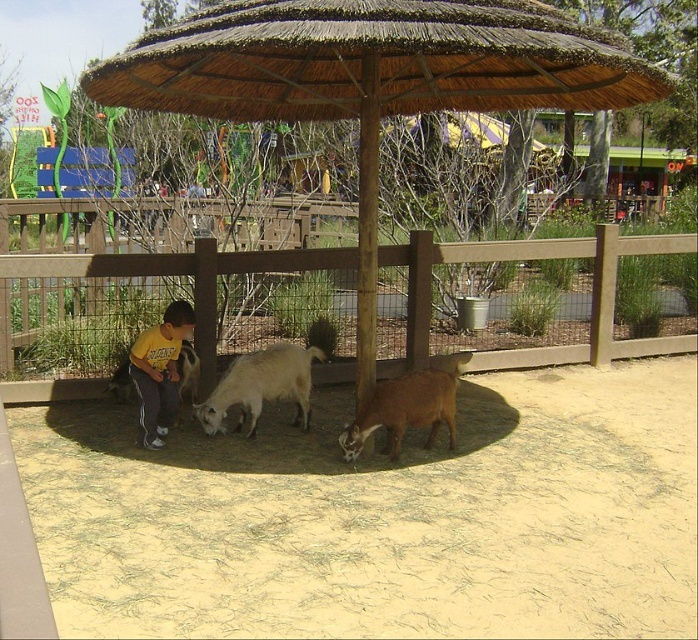}
    \scriptsize 
    \parbox{0.95\linewidth}{
    \texttt{Q : Where is the place animals are caged in the image?}\\
    \texttt{KB : [caged\_animals, AtLocate, zoo]}\\
    \texttt{A : zoo}
    }
  \end{subfigure}
    \caption{BOK-VQA dataset examples}
    \label{fig:appendix-dataset-examples}
\end{figure*}

\subsection{A. Dataset Details}\label{subsec:appendix-dataset}
In this section, we present the details of the dataset, including examples (figure~\ref{fig:appendix-dataset-examples}) and dataset statistics(Table~\ref{tab:appendix-relation-statistics}, Figure~\ref{fig:pieplot1}, Figure~\ref{fig:pieplot2}).

\begin{figure}[h]
    \centering
    \includegraphics[width=\linewidth]{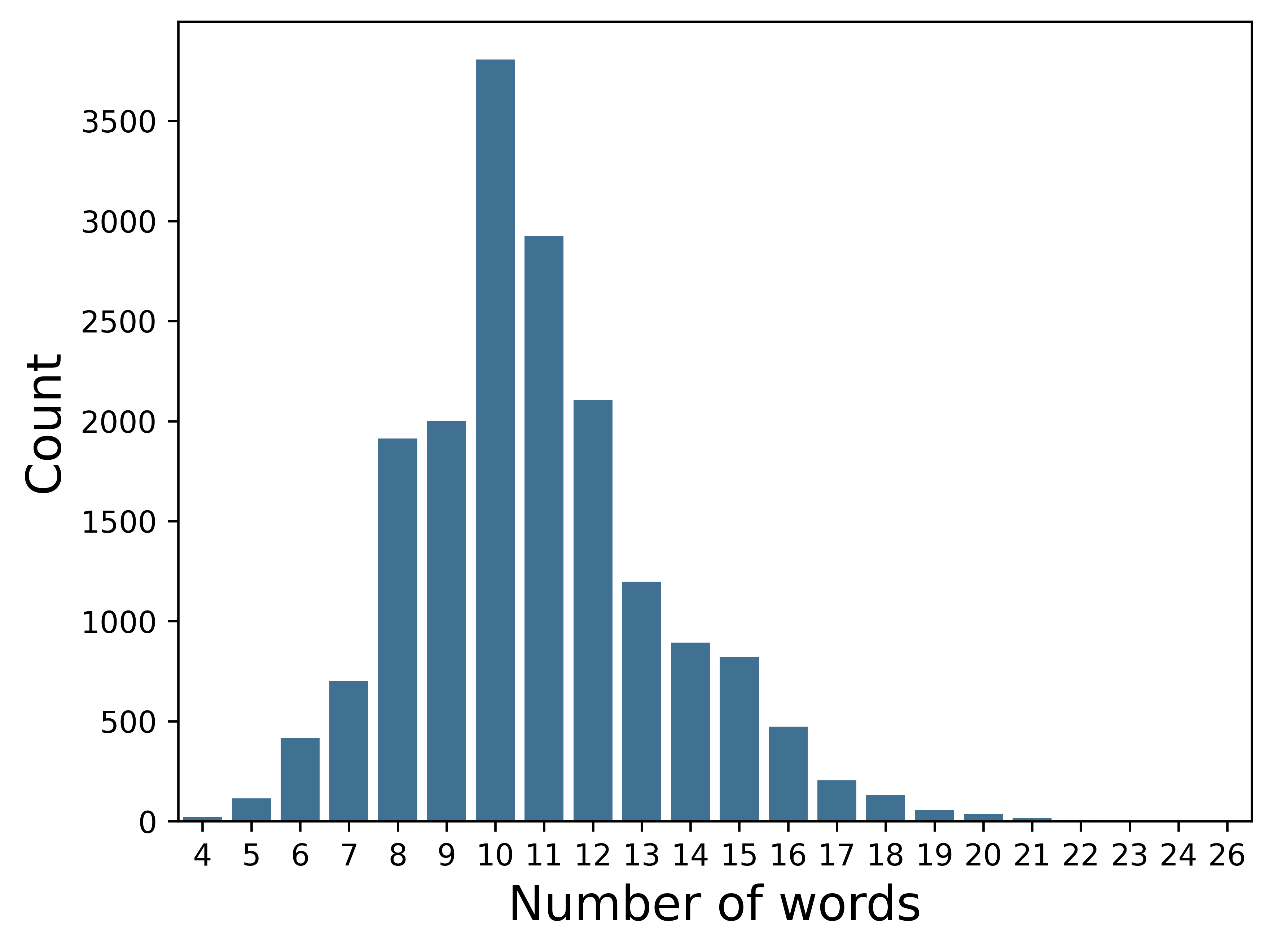}
    \caption{Word count distribution of questions}

\end{figure}

\begin{figure}[h]
    \centering
    \includegraphics[width=\linewidth]{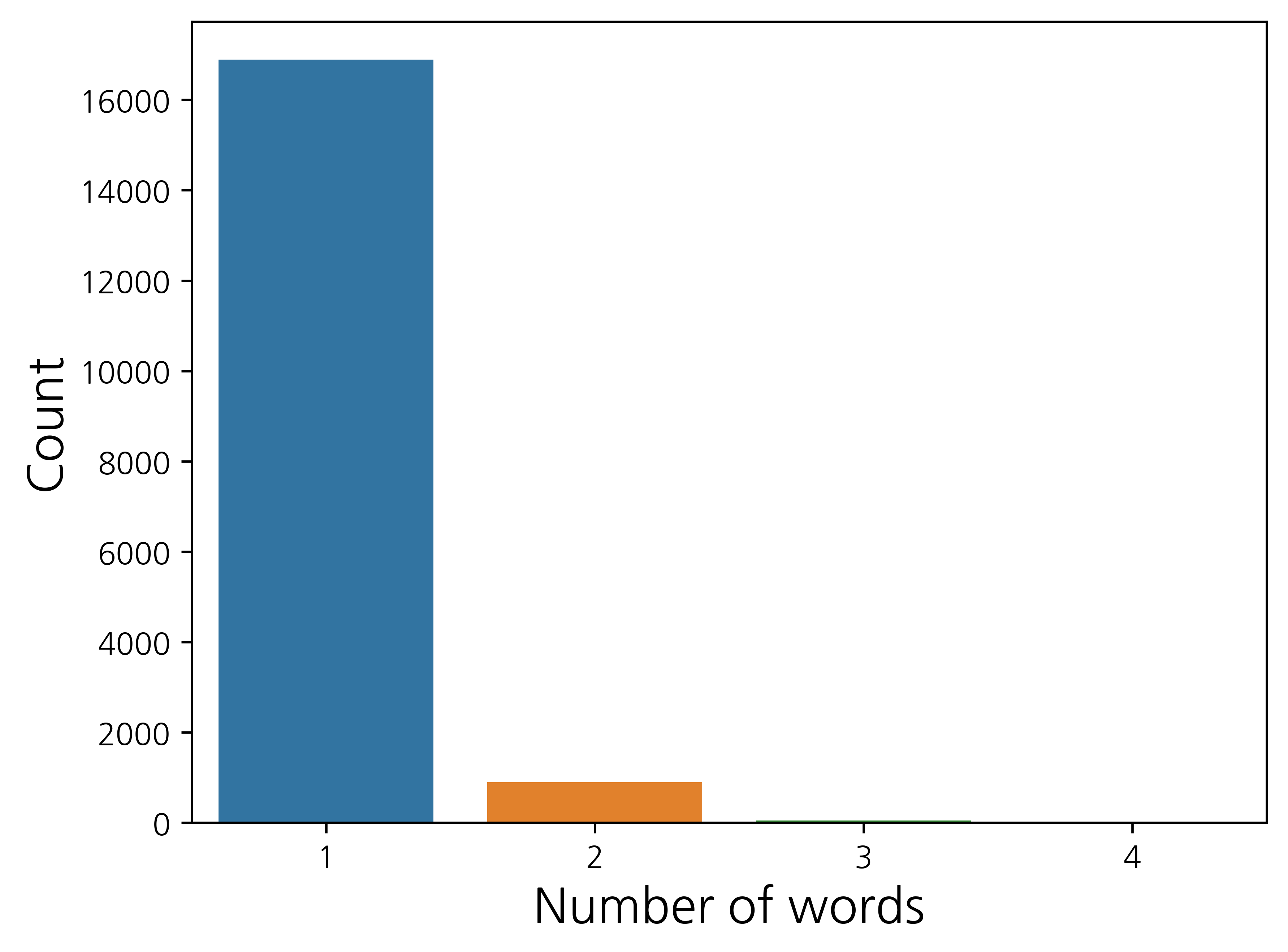}
    \caption{Word count distribution of answers}

\end{figure}

\subsection{B. Triple Prediction Module Details}
\label{sec:TPModule-details}
In the Triple Prediction Module, a heterogeneous linear layer is utilized to predict the Head, Relation, and Tail. This involves having shared parameters as well as specific parameters for the classifiers of the Head, Relation, and Tail. If the entire Triple Prediction Module is denoted as $P_{kb}$,the following equation can be represented as:
$$p^{(h)}=P_h(h_i|v_i, q_i;\theta_s, \theta_h)$$
$$p^{(r)}=P_r(r_i|q_i;\theta_s, \theta_r)$$
$$p^{(t)}=P_t(t_i|q_i;\theta_s, \theta_t)$$
$$P_{kb}(h_i, r_i, t_i|v_i, q_i; \theta_s, \theta_{hrt}) = p^{(h)}\cdot p^{(r)} \cdot p^{(t)}$$

In the model, the parameters $\theta_{s}$ shared by the image and question feature extraction models, along with parameters $\theta_h, \theta_r, \theta_t$ for the classifiers of the Head, Relation, and Tail, respectively, are used. Here, $\theta_{hrt} = (\theta_h, \theta_r, \theta_t)$, where $v_i, q_i$ correspond to the $i$'th image and question, and $h_i, r_i, t_i$ denote the targets for the head, relation, and tail, respectively. Let the optimal parameter set be denoted as $\theta^* = (\theta_s^*, \theta_h^*, \theta_r^*, \theta_t^*)$. The triple classification module can be defined in the form of Maximum-likelihood estimation.

$$\theta^* = \max_{\theta_s, \theta_h, \theta_r, \theta_t} \prod_{i=1}^N p^{(h)}\cdot p^{(r)}\cdot p^{(t)}$$
Let $C_h, C_r, C_t$ represent the number of classes for the Head, Relation, and Tail respectively, and $\mathcal{L}_h, \mathcal{L}_r, \mathcal{L}_t$ denote their respective losses. Then, the loss function $\mathcal{L}_\mathrm{T}$ for the Triple prediction can be constructed as follows:
$$\mathcal{L}_h(\theta_s, \theta_h) = - \frac{1}{N} \sum_{i=1}^{N} \sum_{j=1}^{C_h} h_{i,j} \log p^{(h)}$$
$$\mathcal{L}_r(\theta_s, \theta_r) = - \frac{1}{N} \sum_{i=1}^{N} \sum_{j=1}^{C_r} r_{i,j} \log p^{(r)}$$
$$\mathcal{L}_t(\theta_s, \theta_t) = - \frac{1}{N} \sum_{i=1}^{N} \sum_{j=1}^{C_t} t_{i,j} \log p^{(t)}$$
$$\mathcal{L}_\mathrm{T}(\theta) = \mathcal{L}_h(\theta_s, \theta_h) + \mathcal{L}_r(\theta_s, \theta_r) + \mathcal{L}_t(\theta_s, \theta_t)$$

Firstly, the feature vector $f^{(v)}_i$ of the image $v_i$ is obtained through the ResNet, and the feature vector $f^{(q)}_i$ of the question $q_i$ is derived through the XLM-RoBERTa. Following the extraction of the feature vectors for each modality, the multimodal feature vector $f^{(u)}_i$ is computed through the element-wise multiplication of $f^{(v)}_i$ and $f^{(q)}_i$.
$$f^{(v)}_i = \text{ResNet}(v_i)$$
$$f^{(q)}_i = \text{XLMRoBERTa}(q_i)$$
$$f^{(u)}_i = f^{(v)}_i \otimes f^{(q)}_i$$

We designed the model as follows, taking into account that the Head contains information from both the image and the question, while the Relation and Tail have more critical information in the question than in the image. The multimodal feature vector $u_i$, used for predicting the Head, passes through a linear transformation layer $L_h$. Additionally, $f_{q_i}$, which is loaded with a considerable amount of question information, predicts the Relation and Tail respectively by passing through the linear transformation layers $L_r$ and $L_t$. If the output feature vectors, which have passed through the linear transformation layers $L_h, L_r, L_t$ from image $v_i$ and question $q_i$, are denoted as $f^{(h)}_i, f^{(r)}_i, f^{(t)}_i$ respectively, they can be expressed as follows:
$$f^{(h)}_i = W_h(f^{(u)}_i) + b_h = L_h(f^{(u)}_i)$$
$$f^{(r)}_i = W_r(f^{(q)}_i) + b_r = L_r(f^{(q)}_i)$$
$$f^{(t)}_i = W_t(f^{(q)}_i) + b_t = L_t(f^{(q)}_i)$$

The final prediction values for each class, $\hat{h}_{i}, \hat{r}_{i}, \hat{t}_{i}$, are obtained from the feature vectors of the Head, Relation, and Tail through the Softmax function.
$$\sigma(z_c) = \frac{e^{z_c}}{\sum_{j=1}^{k} e^{z_{j}}}$$
$$\hat{kb}_{i} = \argmax_{c}(\sigma(f^{(kb)}_{i,c})), \text{ where } kb\in\{h, r, t\}$$

\subsection{C. Hyperparamter Details} \label{sec:appendix-hyper}
In our experiment, we utilized the ConvKB knowledge graph embedding model to pretrain 3,797 triples into a size of 256 over 2,000 epochs. To take advantage of both Korean and English in the BOK-VQA dataset, we employed the pretrained multilingual model XLM-RoBERTa. All experiments were conducted over 50 epochs using the AdamW optimizer with a learning rate of 5e-5. The detailed hyperparameter settings are provided in Table~\ref{tab:hyperparameter}.

\begin{table}[h]
\centering
{\begin{minipage}{0.4\textwidth} 
    \small
    \begin{tabular}{l | c } 
    \hline
{component} & {value} \\ 
\hline
 Train,Valid,Test & 60\%,20\%,20\% \\ 
 $W$ (parameters) dim. & 256 \\
 $v_{i}$ (ResNet50) dim. & 768 \\ 
 $f_{i}$ (ResNet50) dim. & 768 \\ 
 $q_{i}$ (RoBERTa) dim. & 768 \\ 
 $h_{i},r_{i},t_{i} $ dim. & 256 \\
 MLP hidden dim. & 768 \\ 
 Dropout & 0.2 \\ 
 Learning rate & 5e-5  \\
 Optimizer & AdamW \\ 
 $\beta_1$, $\beta_2$  & 0.9, 0.99  \\ 
 Epoch for VQA & 50  \\ 
 Iteration for KGE & 50,000  \\ 
 Batch size (VQA)  & 128  \\
 Batch size (KGE)  & 512  \\
 {Random Seed} & {41,42$^{*}$,43,44, and 45}  \\ 

 \hline
  \end{tabular}
\end{minipage}}
  \caption{Hyperparameters.}   \label{tab:hyperparameter}

\end{table}

The learning data was divided into Train/Valid/Test data at a ratio of 60:20:20, respectively. Performance evaluation was conducted by applying a 5-fold cross-validation, and it was calculated using the formula Eq.~\eqref{eq-acc}.

\begin{equation} \label{eq-acc}
  \text{ACC} = \frac{\text{correctly classified number of examples}}{\text{total number of examples}}
\end{equation}{}

\subsection{D. Application of Teacher Forcing and Self-Attention in GEL-VQA}\label{sec:appendix-c}
\begin{figure}[h]
\centering
\includegraphics[width=\columnwidth]{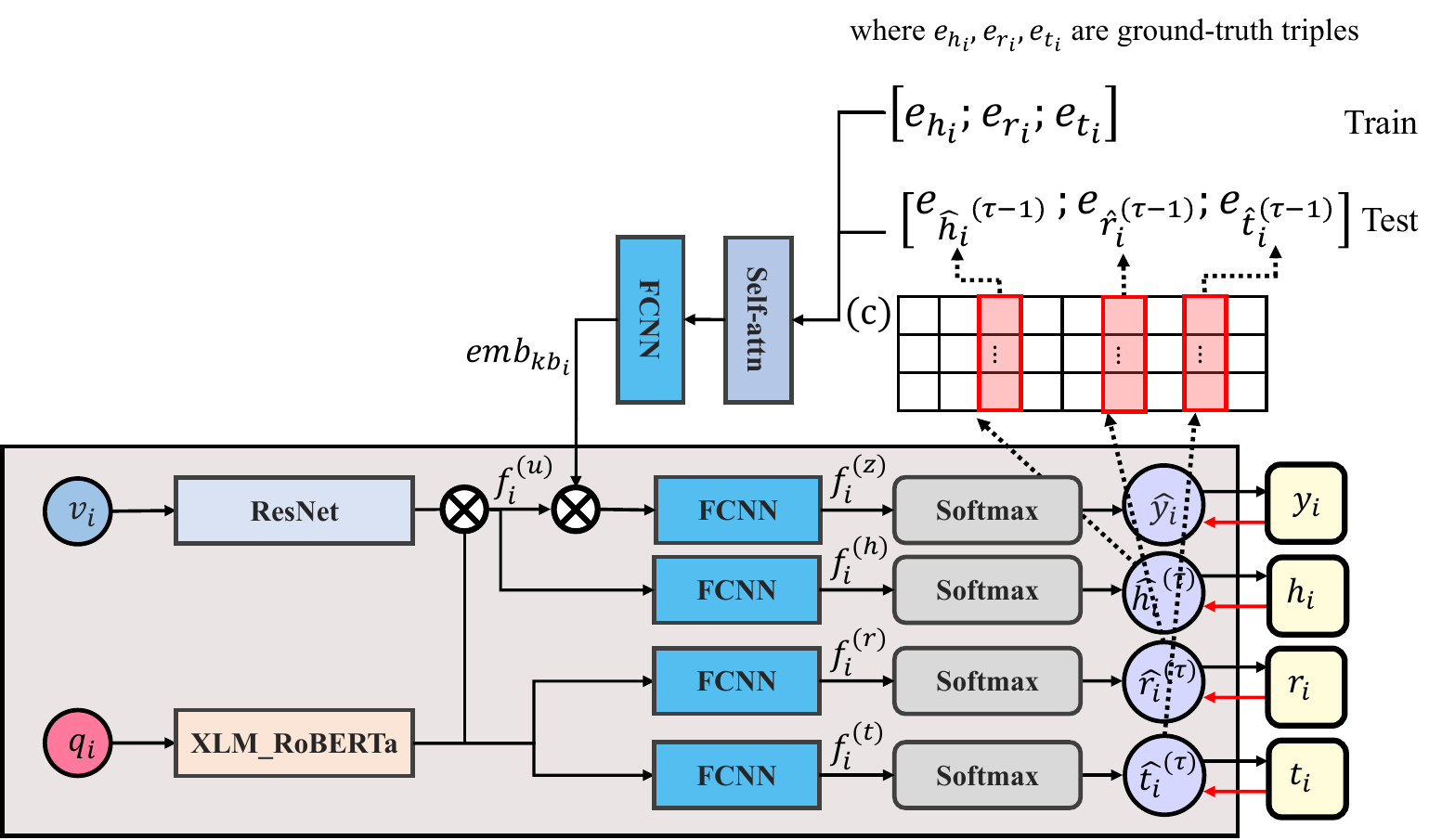}
\caption{Application of teacher forcing and self-attention in \textsc{gel-vqa}}
\end{figure}
The proposed model applies Teacher Forcing to Triple prediction. In the \textsc{gel-vqa} model, the triples predicted at the previous time step were used for model training. However, in the \textsc{gel-vqa}+TF model, the target values of the actual triples are used as inputs. By utilizing Teacher Forcing, we prevent the distortion of learning due to incorrect predictions at previous time steps. Additionally, in the \textsc{gel-vqa} +TF + ATTN model, self-attention is applied to the triple embedding $emb_{kb}$. Through self-attention, the model learns where to focus more among the head, tail, and relation.

\begin{multline}
\mathcal{L}_{\mathrm{VQA}}^{(\tau)}(\theta) =  - \frac{1}{N} \sum_{i=1}^{N} \sum_{j=1}^{C}\\ y_{i, j} \log(P_{vqa}(j | v_i, q_i, \mathcal{E}_i; \theta))
\end{multline}

where $\mathcal{E}_i$ is the set of ground-truth triple $\{h_i, r_i, t_i\}$.

\begin{equation}
\mathcal{L}^{(\tau)}(\theta) =  \mathcal{L}_{\mathrm{VQA}}(\theta) + \mathcal{L}_{\mathrm{T}}(\theta)
\end{equation}

\subsection{E. Attention Visualization of Representative Samples}

\begin{figure}[h]
    \caption*{Sample 1.}
    \includegraphics[width=\linewidth]{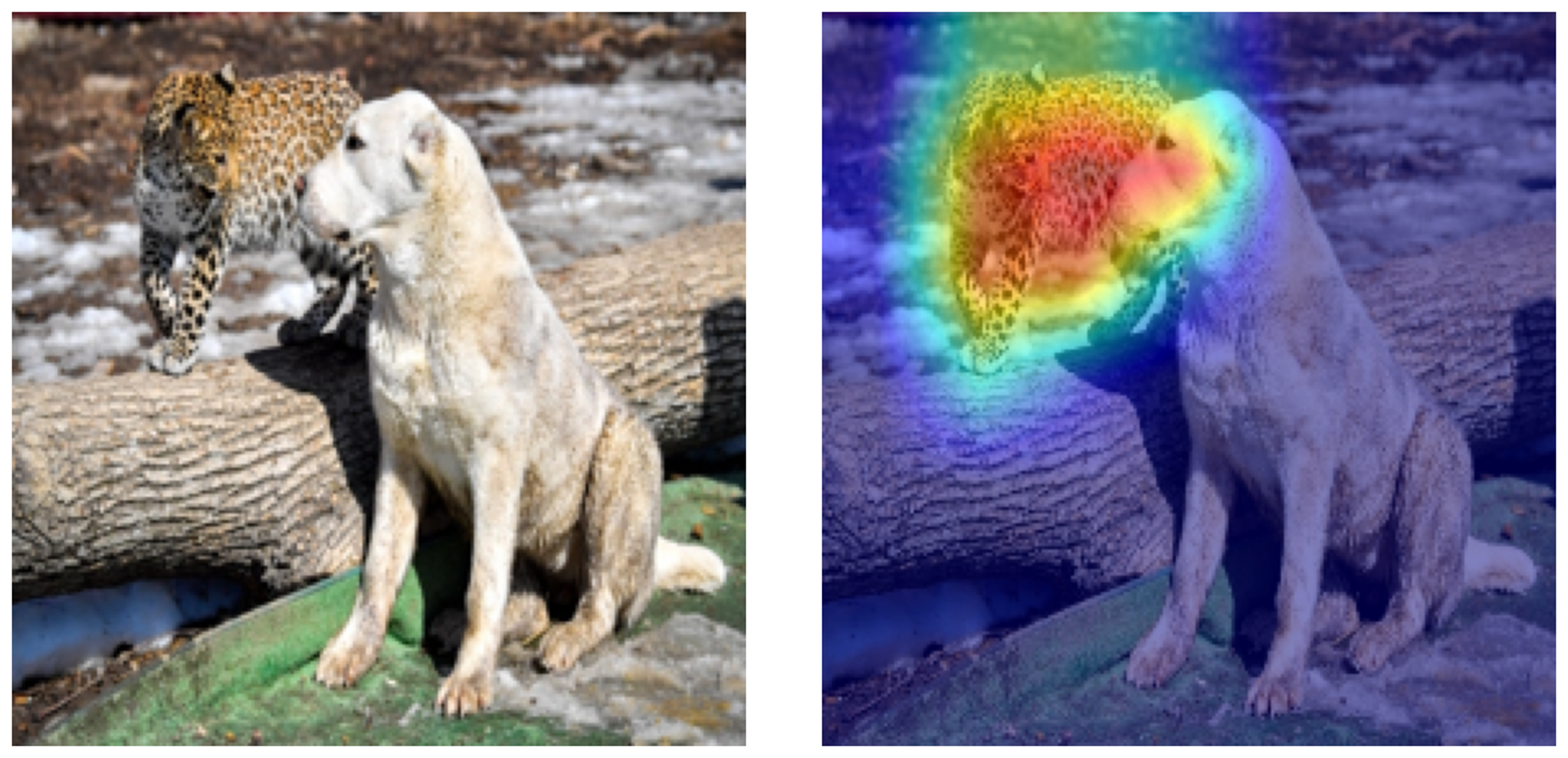}
    \captionsetup{format=plain}
    \caption*{
    \scriptsize question : \texttt{What animals are the symbol of the Democratic Republic of Congo in the image?}\\
    kb : \texttt{[Leopard, Subject, the symbol of the Democratic Republic of Congo]}\\
    Answer : \texttt{Leopard}\\
  }
  
    \includegraphics[width=\linewidth]{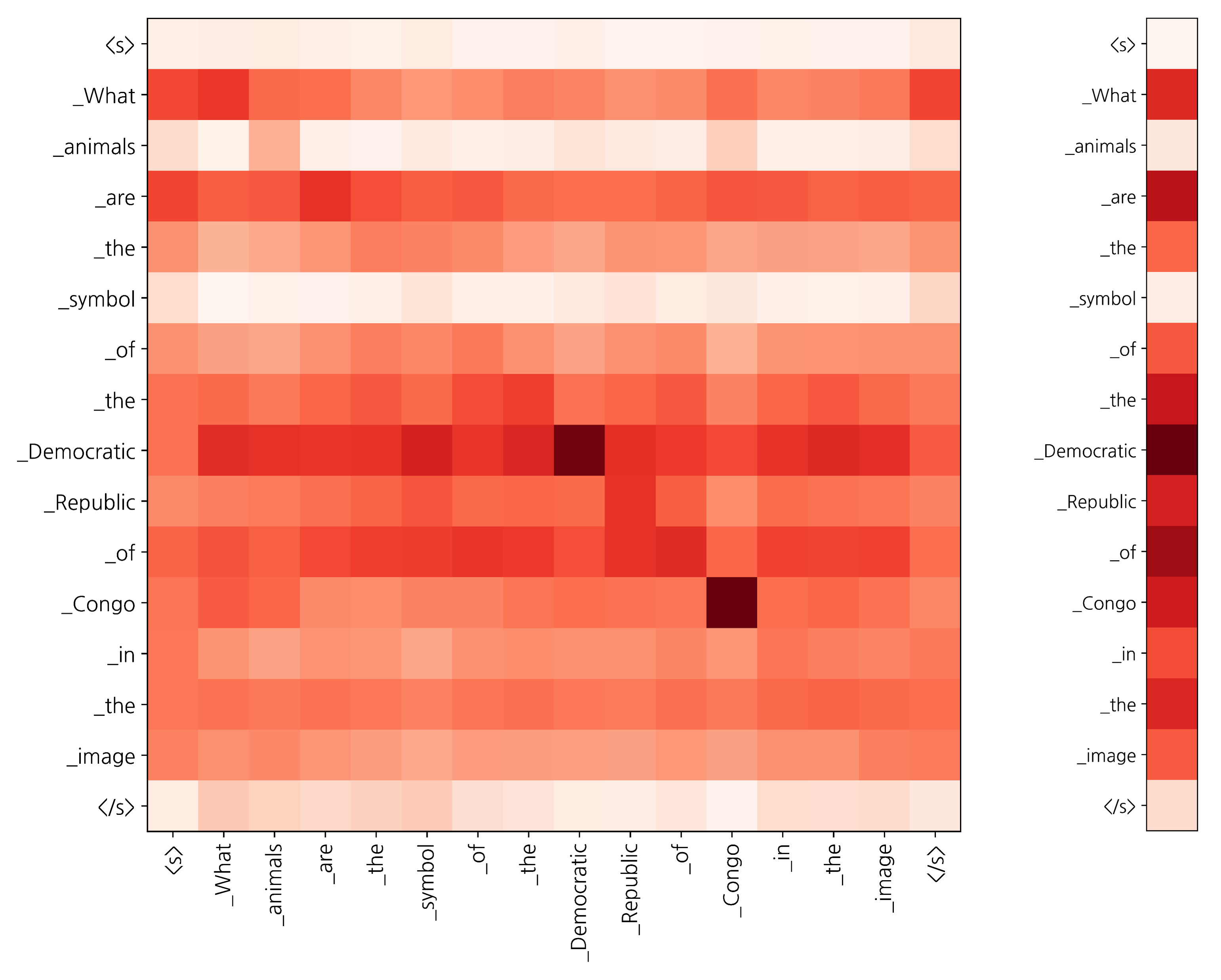}
    \captionsetup{format=plain}
    \caption*{\scriptsize Question Token Attention}
    \includegraphics[width=\linewidth]{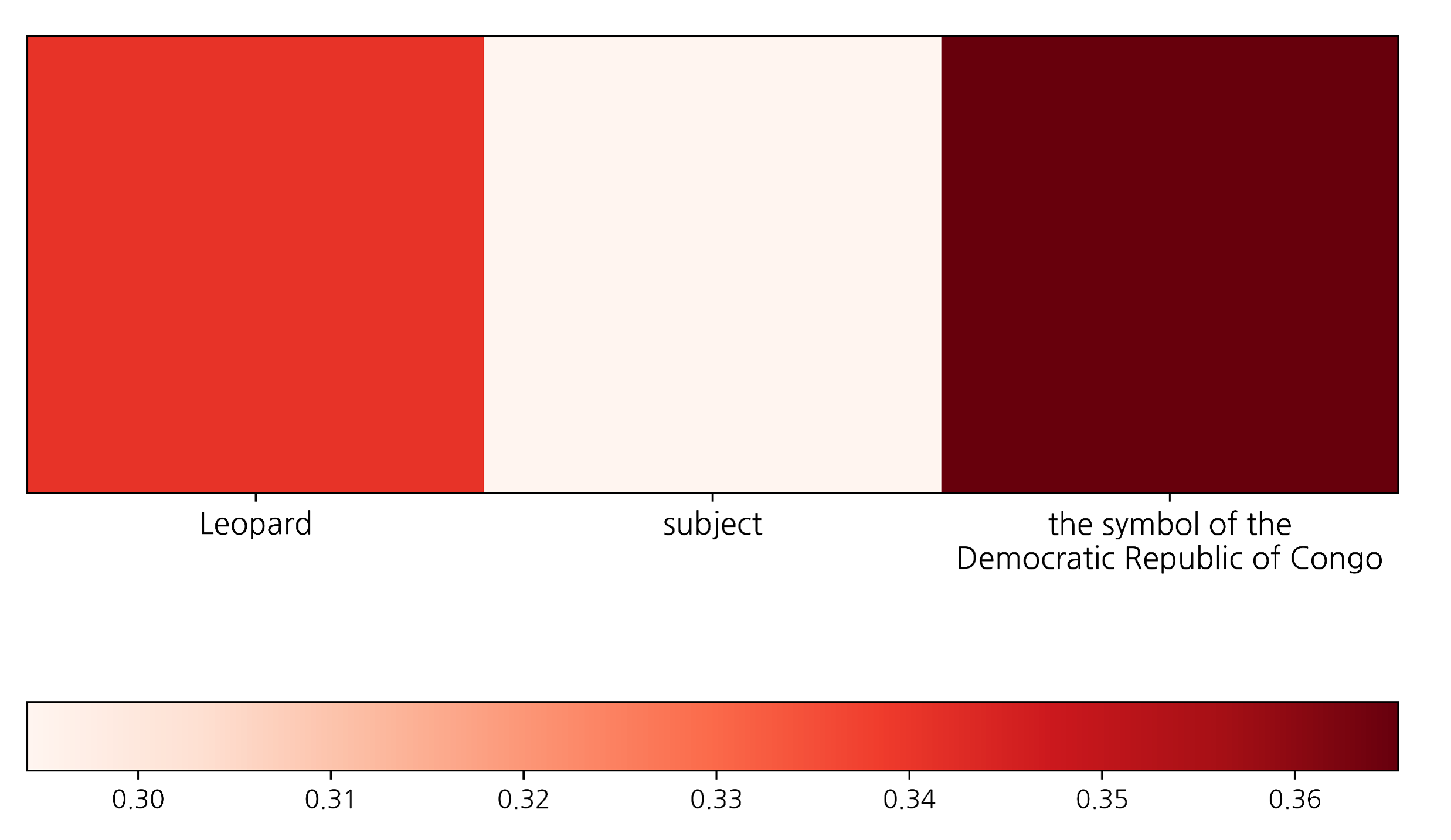}
    \captionsetup{format=plain}
    \captionsetup{skip=0pt}
    \caption*{\scriptsize Triple Attention}
    
\end{figure}

\begin{figure}[h]
    \vspace{1cm}

    \caption*{Sample 2.}
    \includegraphics[width=\linewidth]{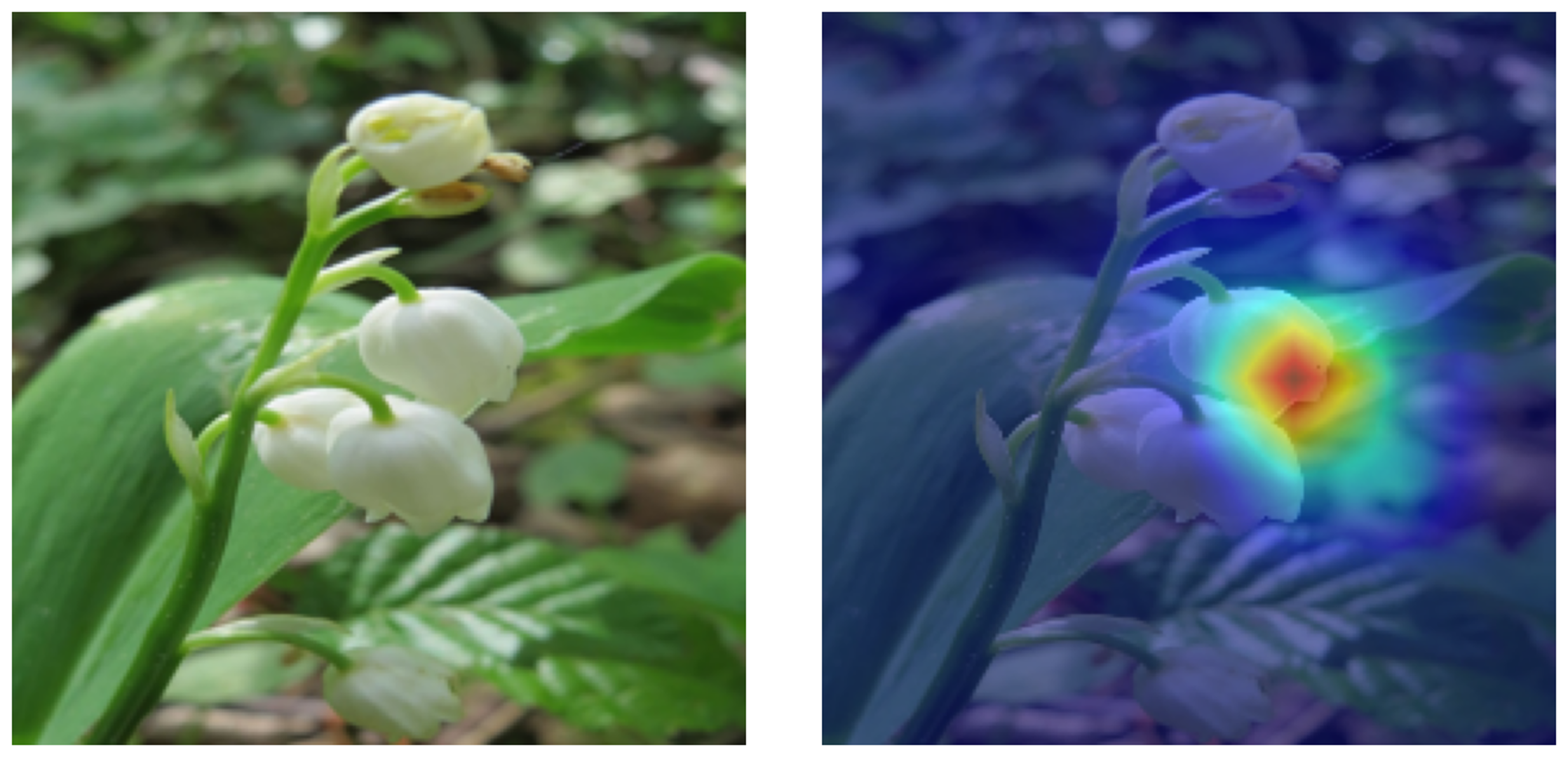}
    \captionsetup{format=plain}
    \caption*{
    \scriptsize question : \texttt{Which plant belongs to the asparagus family in the image?}\\
    kb : \texttt{[lily of the valley, family, asparagaceae]}\\
    Answer : \texttt{lily of the valley}\\
  }
  
    \includegraphics[width=\linewidth]{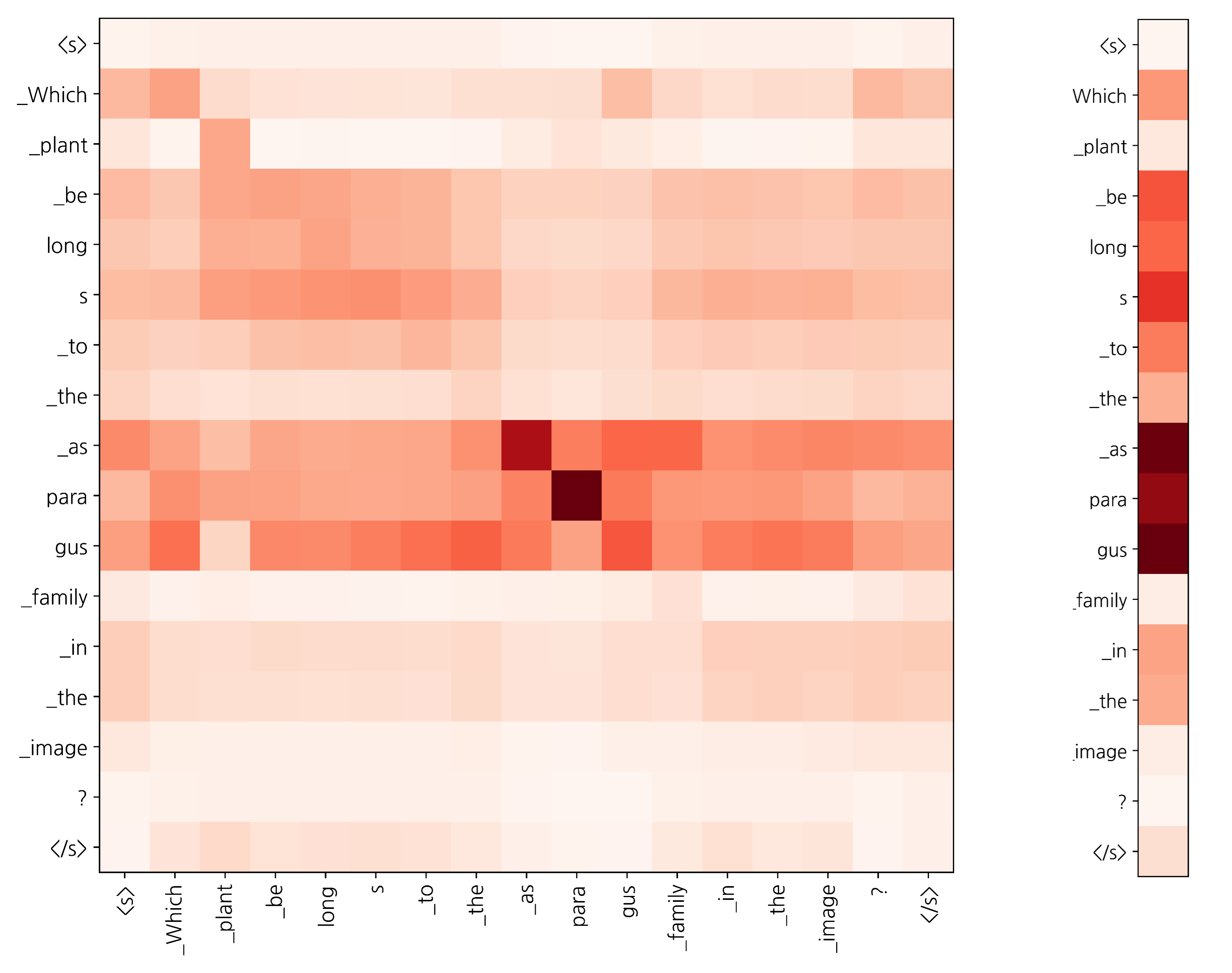}
    \caption*{\scriptsize Question Token Attention}
    \includegraphics[width=\linewidth]{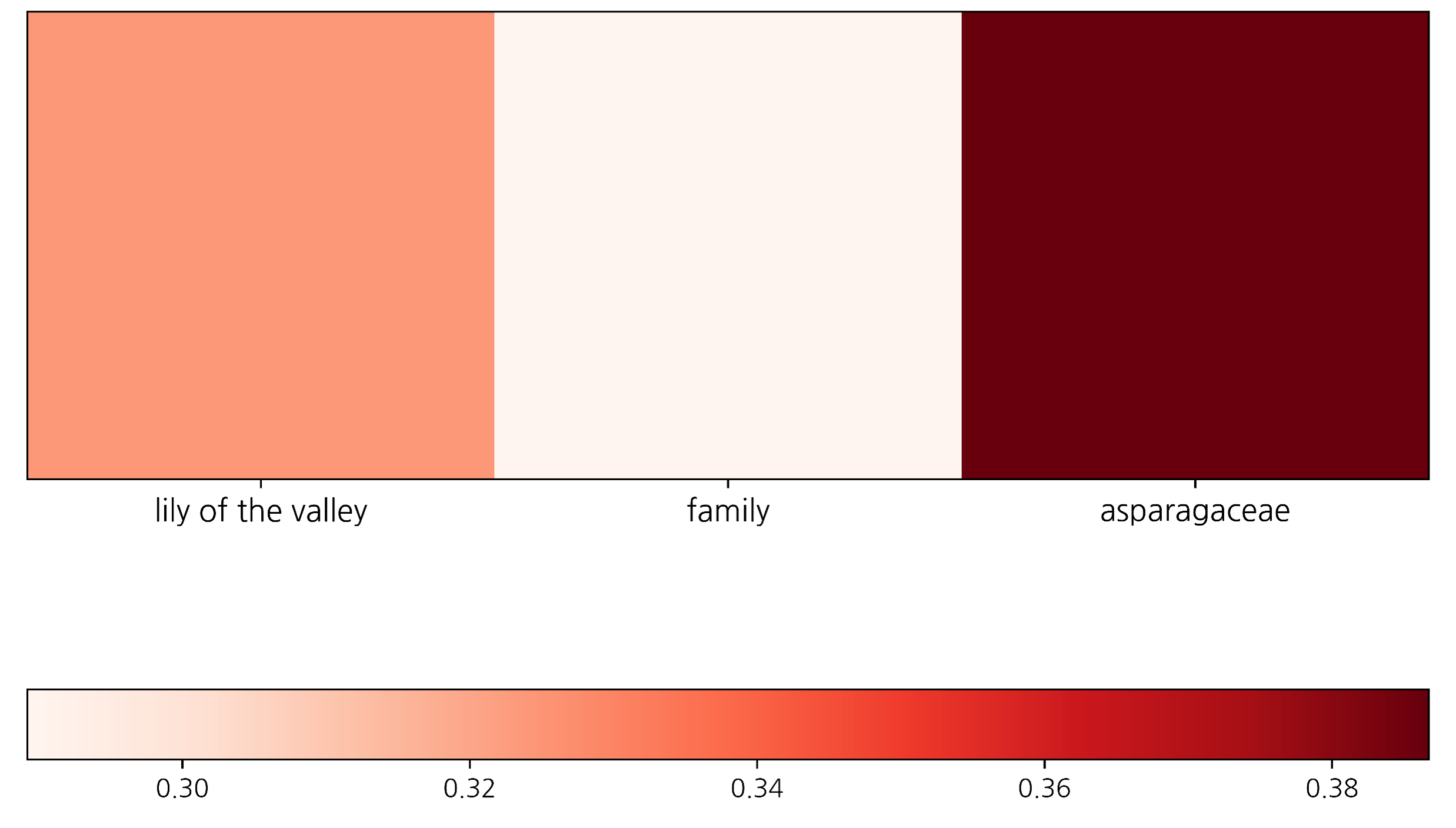}
    \caption*{\scriptsize Triple Attention
    \texttt{ }\\
    \texttt{ }\\}
    
\end{figure}

\begin{figure}[h]
    \caption*{Sample 3.}
    \includegraphics[width=\linewidth]{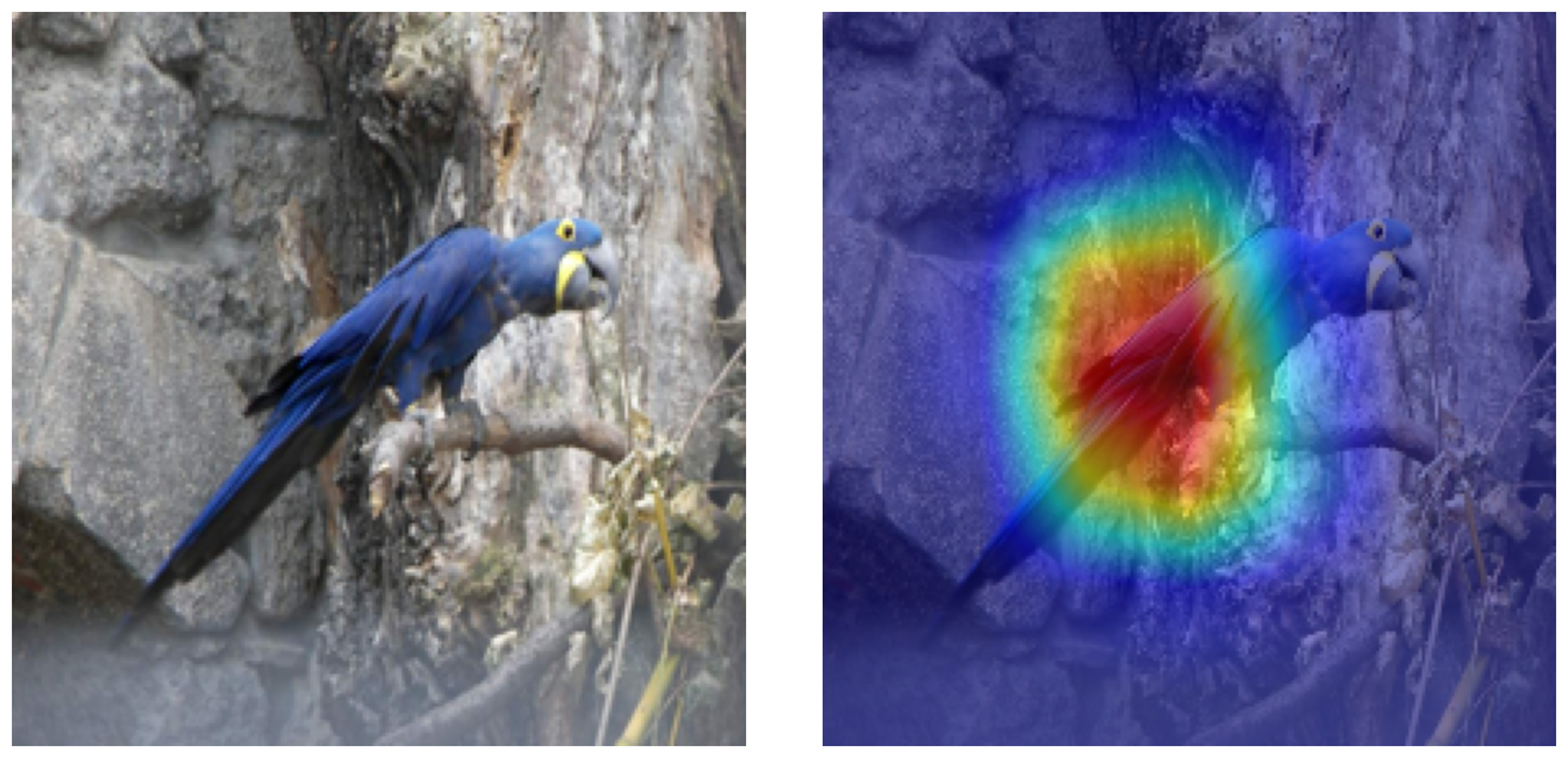}
    \captionsetup{format=plain}
    \caption*{
    \scriptsize question : \texttt{What in the image is related to Central America?}\\
    kb : \texttt{[macaw, RelatedTo, Central America]}\\
    Answer : \texttt{macaw}\\
    \texttt{ }\\
  }
  
    \includegraphics[width=\linewidth]{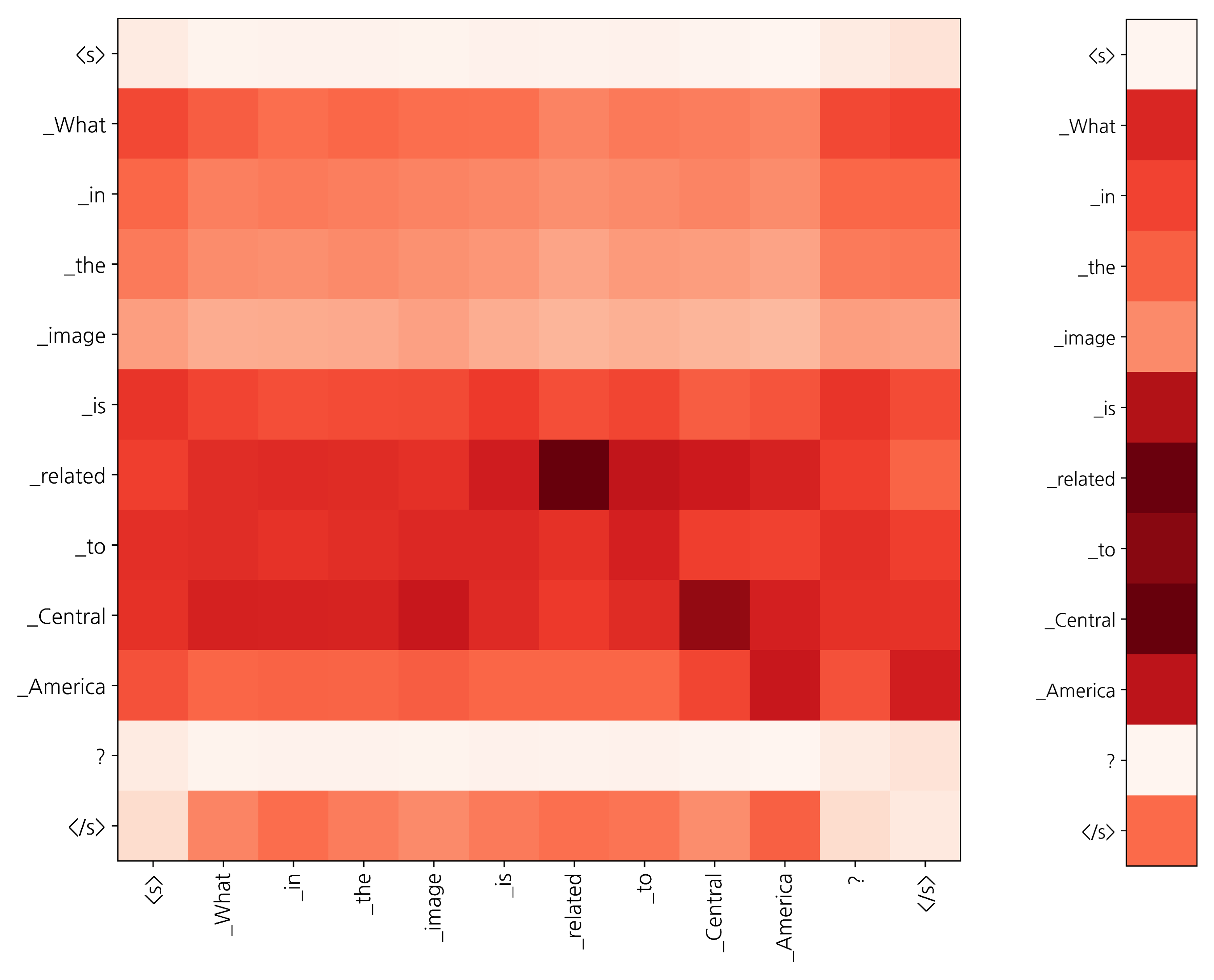}
    \caption*{\scriptsize Question Token Attention}
    \includegraphics[width=\linewidth]{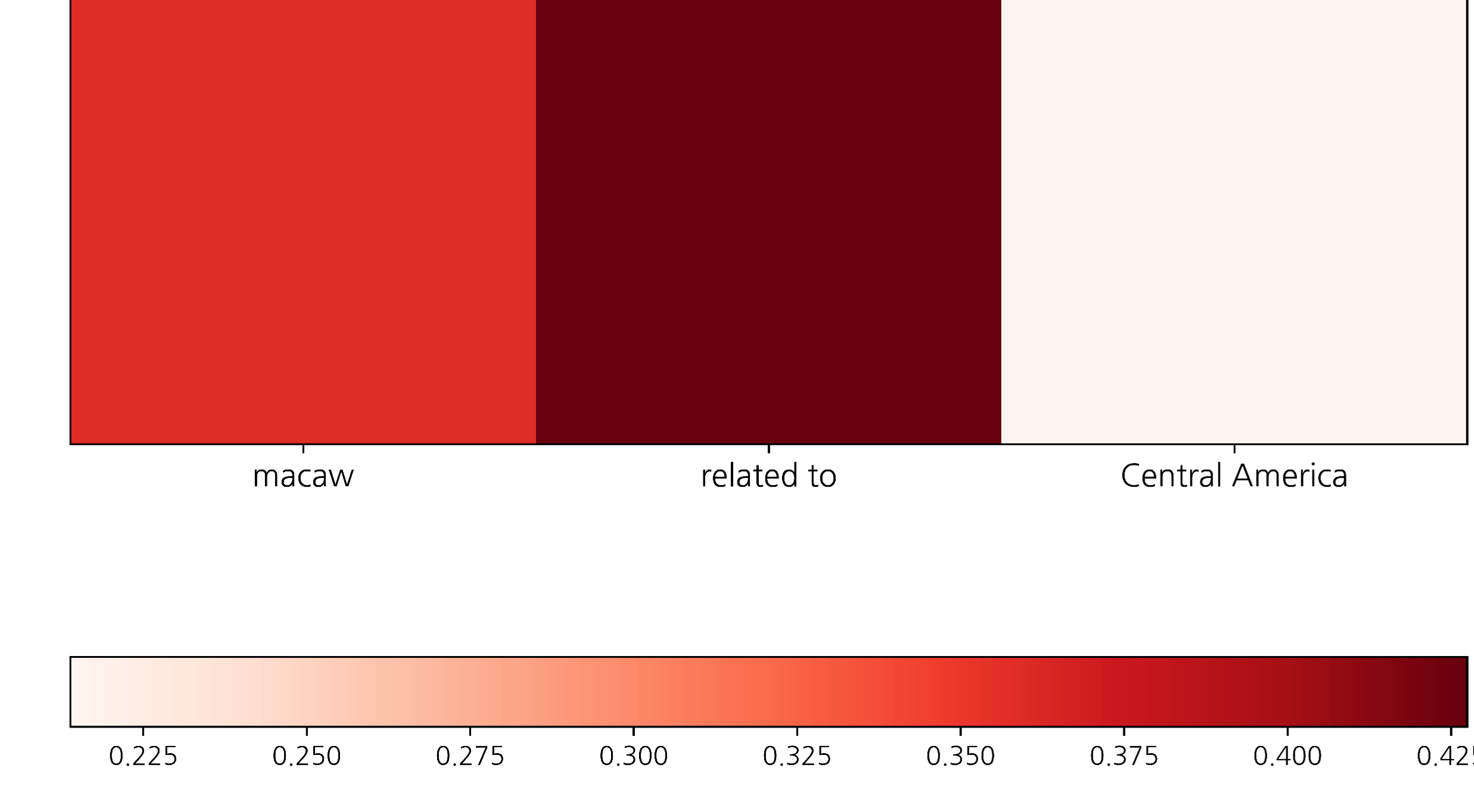}
    \caption*{\scriptsize Triple Attention}
    
\end{figure}
\begin{figure}[h]
    \caption*{Sample 4.}
    \includegraphics[width=\linewidth]{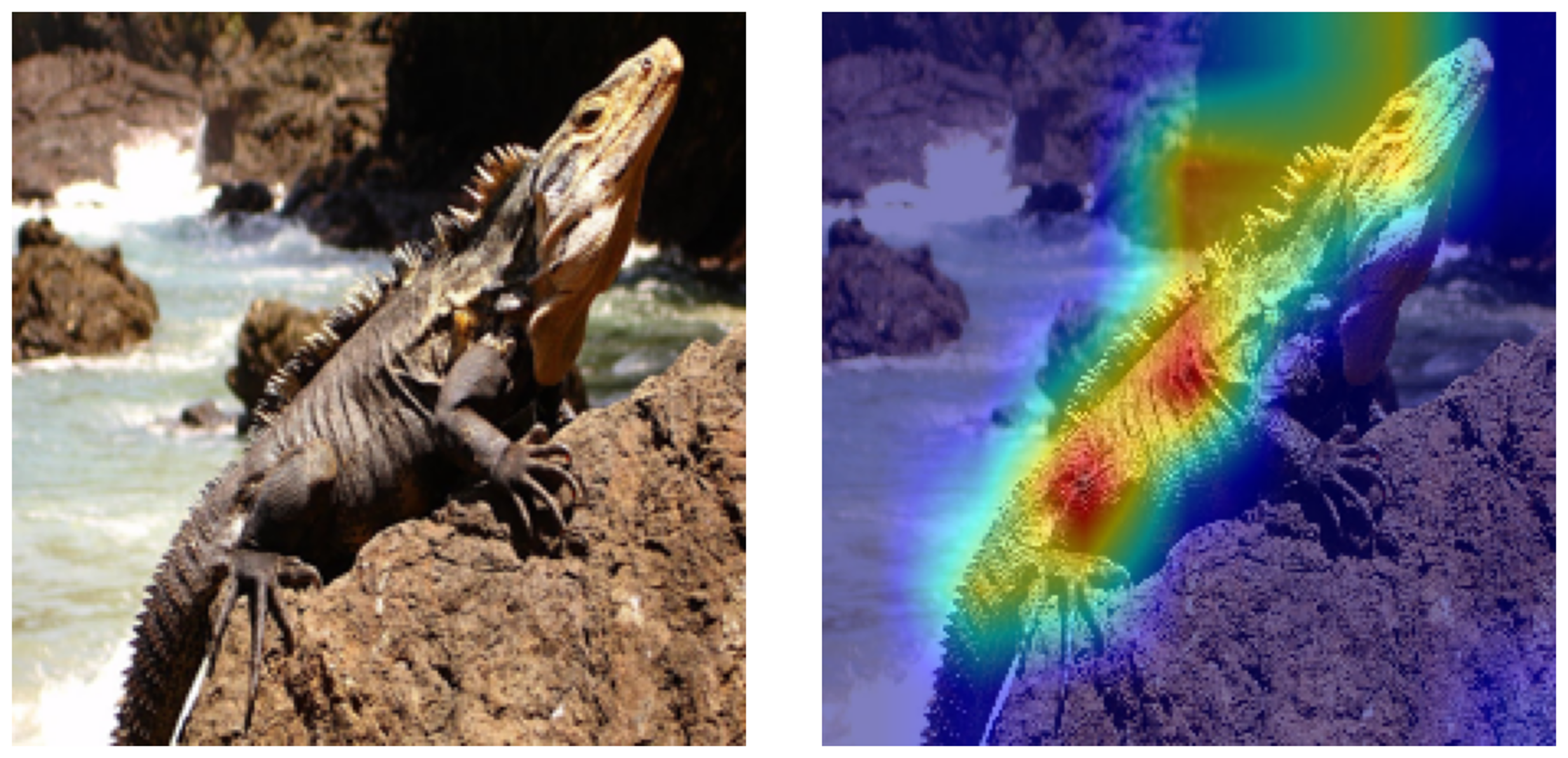}
    \captionsetup{format=plain}
    \caption*{
    \scriptsize question : \texttt{What category does the iguana in the image belong to?}\\
    kb : \texttt{[iguana, order, squamata]}\\
    Answer : \texttt{squamata}\\
  }
  
    \includegraphics[width=\linewidth]{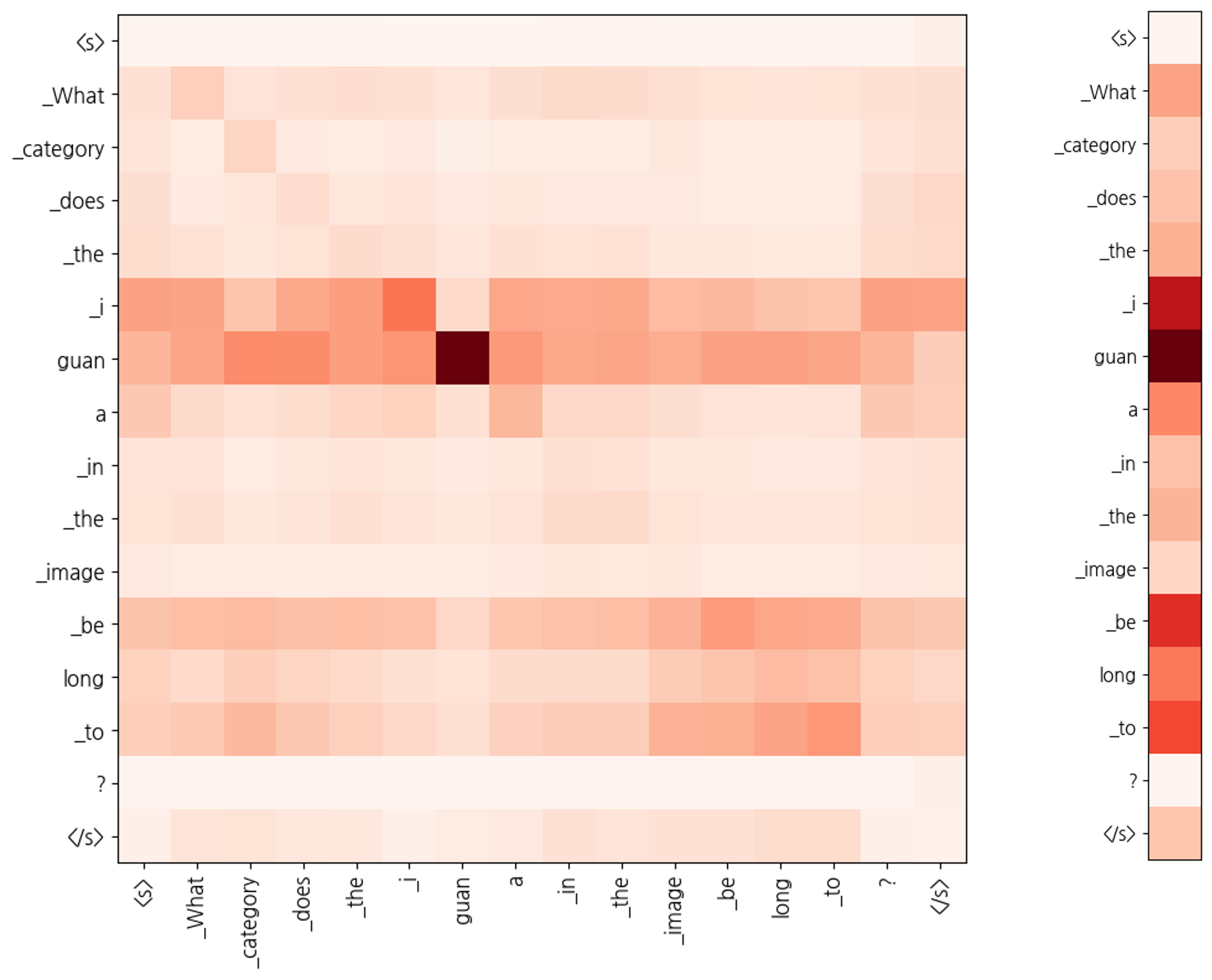}
    \caption*{\scriptsize Question Token Attention}
    \includegraphics[width=\linewidth]{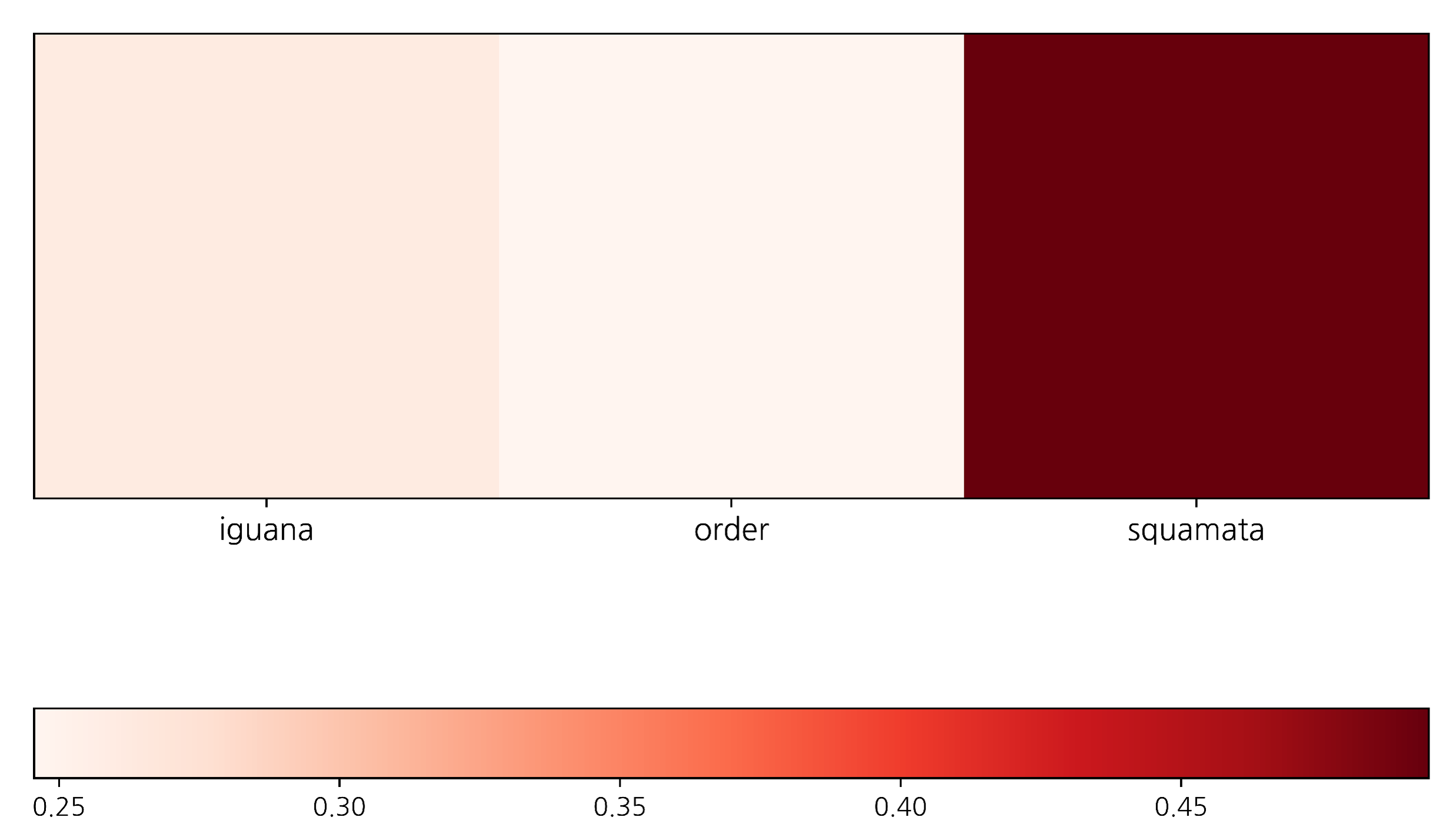}
    \caption*{\scriptsize Triple Attention}
    
\end{figure}

\end{document}